\documentclass[11pt,letterpaper]{article}

\usepackage{modalens-preprint}

\usepackage[utf8]{inputenc}
\usepackage[T1]{fontenc}
\usepackage[hyphens]{url}
\usepackage{booktabs}
\usepackage{array}
\usepackage{adjustbox}
\usepackage{float}
\usepackage{amsmath}
\usepackage{amssymb}
\usepackage[table]{xcolor}
\usepackage{tikz}
\usepackage{natbib}
\usepackage[most]{tcolorbox}
\newif\ifreviewmarks \reviewmarksfalse   
\newcommand{\newmarkfill}{yellow!30}
\makeatletter
\newcommand{\nw@word}[1]{{\setlength{\fboxsep}{0.6pt}\colorbox{\newmarkfill}{\vphantom{Ayg}#1}}}
\def\nw@scan#1 #2\nw@stop{%
  \nw@word{#1}%
  \ifx\relax#2\relax\else\space\nw@scan#2\nw@stop\fi}
\DeclareRobustCommand{\new}[1]{\ifreviewmarks\nw@scan#1 \nw@stop\else#1\fi}
\makeatother
\newenvironment{newpar}[1][]%
  {\ifreviewmarks
     \tcolorbox[breakable,enhanced jigsaw,colback=\newmarkfill,colframe=\newmarkfill,
       boxrule=0pt,arc=0pt,left=2pt,right=2pt,top=3pt,bottom=3pt,
       before skip=\medskipamount,after skip=\medskipamount,parbox=false,#1]%
   \fi}%
  {\ifreviewmarks\endtcolorbox\fi}

\definecolor{aumark}{rgb}{0.00,0.32,0.75}

\newcommand{\modalens}{\textsc{ModaLens}}

\usetikzlibrary{arrows.meta,positioning,calc,fit,backgrounds}

\definecolor{mlDeep}{HTML}{0B3D66}
\definecolor{mlMid}{HTML}{2A72B5}
\definecolor{mlLite}{HTML}{7FB2DC}
\definecolor{mlWarm}{HTML}{B4462F}
\definecolor{mlGrid}{HTML}{DFE4EA}
\definecolor{mlMute}{HTML}{6E6E6E}

\title{\modalens: Measuring Image Sensitivity in Report-Conditioned Medical VLMs}

\newcommand{\aff}[1]{\textsuperscript{#1}}
\author{%
Sebasti\'an Andr\'es Cajas Ord\'o\~nez\aff{1},
Maximin Lange\aff{2,1},
Quang Bui\aff{3},
Anqi Peter Li\aff{4},
Felipe Ocampo Osorio\aff{1},
Rafi Al Attrach\aff{1},
Kushul Reddy Palakala\aff{5},
Sahil Kapadia\aff{6},
Zakaria Laouabdia Sellami\aff{7},
Xinyue Zhang\aff{2},
Ashley Zhang\aff{8},
Leo Anthony Celi\aff{1,9}\\[0.7em]
{\small
\aff{1}MIT Critical Data, Massachusetts Institute of Technology\quad
\aff{2}King's College London\quad
\aff{3}American International School Vienna\quad
\aff{4}Substrate Labs\quad
\aff{5}School of Computing, University of North Florida\quad
\aff{6}Department of Neuroscience, University of North Carolina at Chapel Hill\quad
\aff{7}Motork\quad
\aff{8}Collingwood School\quad
\aff{9}Beth Israel Deaconess Medical Center}%
}
\date{}
\begin{document}
\maketitle


\begin{abstract}
A radiology report can already answer a clinical question, so it is hard to tell whether a
vision-language model also uses the image. \modalens, a paired image-swap audit, measures how report
availability changes image sensitivity: MedGemma-27B on 3{,}199 paired MIMIC-CXR cases from 293
patients, all 14 questions per case (13 finding-specific and one composite), each image replaced
by one from another study,
usually of the same patient, with question and report fixed. \new{Under an explicit answer instruction, the model's generated
answer changes on 4.26\% of trials with the report and 20.94\% without it, a paired increase of
16.7 points (patient-clustered 95\% CI [15.6, 17.7]), so report availability reduces image-swap
sensitivity under this protocol; the original prompt with a lowercase first-token readout gives
4.70\% against 17.07\%, and substitutions also move continuous answer scores where the binary
prediction does not change. The labels are derived from reports, which limits conclusions about
visual correctness; the direction replicates in two further model lineages.}
Code, the exact prompts and a run record for every number are at \url{https://github.com/criticaldata/MODALENS}.
\end{abstract}

\begin{figure}[H]
\centering
\resizebox{\textwidth}{!}{\begin{tikzpicture}[
  font=\sffamily\footnotesize,
  badge/.style={fill=mlDeep,text=white,font=\sffamily\small\bfseries,minimum size=.46cm,
                inner sep=1pt,rounded corners=1.5pt,anchor=west},
  hd/.style={mlDeep,font=\sffamily\small\bfseries,anchor=west},
  box/.style={draw=mlGrid,line width=.8pt,rounded corners=2pt,inner sep=4pt,align=center,
              minimum height=.72cm},
  lbl/.style={mlMute},
  val/.style={mlDeep,font=\sffamily\small\bfseries,anchor=west}]

\node[badge] (bA) at (0.10,3.40) {A};
\node[hd,right=.12cm of bA] {One case, two inputs};
\node[lbl,anchor=east] at (1.25,2.42) {concordant};
\node[lbl,anchor=east,mlWarm] at (1.25,1.12) {discordant};
\node[box,fill=mlMid!18,draw=mlMid,minimum width=1.95cm] at (2.35,2.42) {own image};
\node[box,fill=mlWarm!18,draw=mlWarm,minimum width=1.95cm] at (2.35,1.12) {\textbf{other} image};
\node[box,fill=white,minimum width=1.30cm,text=mlMute] at (4.20,2.42) {report};
\node[box,fill=white,minimum width=1.30cm,text=mlMute] at (4.20,1.12) {report};
\draw[mlWarm,line width=.8pt,dashed,-{Stealth[length=5pt,width=4pt]}] (2.35,2.00) -- (2.35,1.54);
\node[lbl,anchor=west] at (0.10,0.32) {only the image is swapped};

\node[badge] (bB) at (5.35,3.40) {B};
\node[hd,right=.12cm of bB] {Block the report};
\draw[draw=mlGrid,line width=.8pt] (5.85,0.78) rectangle (7.65,2.90);
\foreach \y in {0.98,1.18,...,2.79}{\draw[mlGrid,line width=.3pt] (5.88,\y)--(7.62,\y);}
\node[lbl,rotate=90,anchor=south] at (5.71,1.84) {decoder layer};
\begin{scope}[shift={(5.92,0.83)},x={(10.5cm,0)},y={(0,0.033115cm)}]
  \draw[mlGrid,line width=.6pt,dashed] (0.0150,0)--(0.0150,61);
  \draw[mlGrid,line width=.6pt,dashed] (0.1200,0)--(0.1200,61);
  \draw[mlWarm,line width=1.3pt] plot coordinates{(0.0775,0)(0.0875,4)(0.0925,8)(0.1075,12)(0.1100,16)(0.1225,20)(0.1400,22)(0.1325,23)(0.0700,24)(0.0650,25)(0.0525,26)(0.0375,27)(0.0200,28)(0.0225,32)(0.0125,36)(0.0150,40)(0.0150,44)(0.0225,48)(0.0175,52)(0.0175,56)(0.0150,60)};
\end{scope}
\node[lbl,anchor=west] at (7.77,0.85) {input};
\node[lbl,anchor=west,mlWarm] at (7.77,1.63) {layer 24};
\node[lbl,anchor=west] at (7.77,2.85) {output};
\node[lbl,anchor=west] at (5.60,0.32) {flips restored $\rightarrow$};

\node[badge] (bC) at (9.25,3.40) {C};
\node[hd,right=.12cm of bC] {Does the answer move?};
\node[lbl,anchor=west] at (9.25,2.84) {with the report};
\fill[mlMid] (9.25,2.27) rectangle (9.99,2.57);
\node[val] at (10.11,2.42) {4.7\%};
\node[lbl,anchor=west] at (9.25,1.54) {without the report};
\fill[mlMid] (9.25,0.97) rectangle (11.95,1.27);
\node[val] at (12.07,1.12) {17.1\%};
\draw[mlGrid,line width=.8pt] (9.25,0.70)--(13.50,0.70);
\node[lbl,anchor=west] at (9.25,0.32) {answer-flip rate, same 44{,}786 trials};

\end{tikzpicture}}
\caption{\textbf{(A)} One case rendered twice: own image (concordant) or another study's
(discordant), report and question fixed. \new{\textbf{(B)} Report-token attention knockout, cumulative from the plotted layer upward, 400
one-question cases: the flip rate rises toward the no-report rate when the block starts shallow, overshoots it at
layer 22, and is at baseline from layer 32. Blocking from early layers raises image-swap sensitivity and blocking from later layers does
not; this measures when direct access to report-token positions matters, not where report
information stops being used
(Appendix~\ref{app:neg}, Figure~\ref{fig:knockout}).} \textbf{(C)} All-14 flip rate, report present and
absent.}
\label{fig:abstract}
\end{figure}
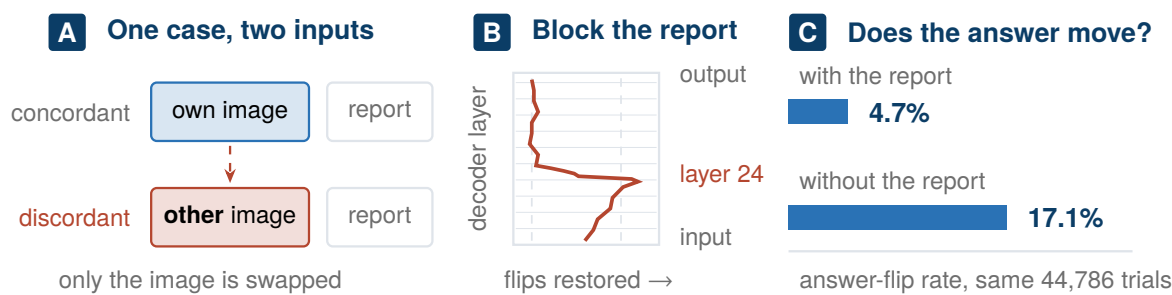

\section{Introduction}
A radiology report can state an answer the image does not support: it may describe a prior study,
be wrong, or name a finding that has since resolved. A model that follows the report is then
correct only when the two agree, and a forced-choice output gives no signal when they do not.
Whether that is a failure depends on the task; what can be measured is how much the output moves
when the image changes and the report stays fixed. That is the line a per-modality failure
framework for clinical multimodal models draws between loud failures, which the system flags, and
silent ones~\citep{bui2026loudsilent}. Audits of this kind are one component of the broader
engineering agenda for humble, accountable clinical decision support that
\citet{arslan2026bodhi} set out.

Answering from text without looking is well documented in general
VQA~\citep{goyal2017making, agrawal2018dont}, persists in multimodal LLMs that miss visually
obvious detail~\citep{tong2024eyes} and hallucinate objects~\citep{rohrbach2018object,li2023pope,leng2024vcd},
and is an explicit debiasing target in medical VQA~\citep{wan2025decoct}. The closest work
is~\citet{lotfinia2026notalways}, which scores image dependence apart from accuracy with target and
control occlusions and same-label image swaps across nine systems, prompting with the finding
question and no report. \modalens{} instead makes report availability the manipulated factor, runs
the same case twice on the same substitution, and reports the paired flip rate and margin change
(Figure~\ref{fig:abstract}, the visual summary of the study).

We estimate how report availability changes image-swap sensitivity within the same cohort
(Table~\ref{tab:designs}); the limitations of that estimate are in Section~\ref{sec:limits}.

\section{Methods}
\modalens{} measures \emph{counterfactual answer sensitivity}: every case is rendered twice, identical
in every token except the image, the \emph{concordant} input with the case's own image and the
\emph{discordant} input with another study's. The whole image is substituted, so the quantity
measured is sensitivity to the image intervention, not grounding of the queried finding.

\textbf{Cohort and units.} MIMIC-CXR test split (Table~\ref{tab:cohort}); a case is one frontal
image of a source study with its own substitute. The estimand is the average over trials; weighting
studies or patients equally, dropping the cross-patient pairs or those with a blank view field, or
keeping only pairs whose known views match gives paired differences between 10.6 and 14.3 points
(Appendix~\ref{app:cohort}).

\begin{table}[H]
\centering\small
\setlength{\tabcolsep}{4pt}
\adjustbox{max width=\textwidth}{%
\begin{tabular}{lrl}
\toprule
Level & Count & Note \\
\midrule
Patients & 293 & resampling unit of the intervals \\
Source studies & 2{,}816 & 344 contribute more than one case \\
Cases (one frontal image, one substitute) & 3{,}199 & 3{,}180 within patient, 19 across patients \\
Substitute studies / images & 1{,}953 / 2{,}093 & reused up to 8 / 6 times; 376 reciprocal pairs \\
Questions per case & 14 & 13 finding-specific plus one composite \\
Paired trials & 44{,}786 & 9{,}347 with the asked finding's label changed \\
\bottomrule
\end{tabular}}
\caption{Cohort accounting; pair composition in Appendix Table~\ref{tab:pairing}.}
\label{tab:cohort}
\end{table}

\textbf{Two question designs.} \emph{All 14 questions} (primary): on every case, 13
finding-specific questions, one per CheXpert finding, plus one composite question asking whether
any acute cardiopulmonary finding is shown (Table~\ref{tab:designs}). The composite question is
scored against an any-finding target, \texttt{yes} iff any of the 13 finding-specific targets is
\texttt{yes}; the dataset's own No Finding label never enters (Appendix Table~\ref{tab:questions}).
\emph{One question per case} (secondary): only the case's own primary labelled finding, or the
composite question with target \texttt{no} on cases with no positive finding (Section~\ref{sec:res2}, Appendix
Table~\ref{tab:ablate}); a label-conditioned sample, not a finding-detection task.

\begin{table}[!htb]
\centering\small
\setlength{\tabcolsep}{4pt}
\adjustbox{max width=\textwidth}{%
\begin{tabular}{lllrlll}
\toprule
Questions per case & Report & Order & Trials & Flip rate & Paired $\Delta$, points & Ratio \\
\midrule
All 14 & present & image first & 44{,}786 & 4.70\% [4.35, 5.06] & reference & reference \\
All 14 & absent  & image first & 44{,}786 & 17.07\% [15.78, 18.39] & $+12.4$ [11.2, 13.6] & 3.6 [3.4, 3.9] \\
\new{All 14} & \new{present} & \new{text first} & \new{44{,}786} & \new{4.91\% [4.62, 5.20]} & \new{$+0.2$ [$-0.1$, 0.5]} & \new{1.0 [1.0, 1.1]} \\
\new{All 14} & \new{absent}  & \new{text first} & \new{44{,}786} & \new{12.75\% [11.76, 13.76]} & \new{$-4.3$ [$-5.4$, $-3.3$]} & \new{0.7 [0.7, 0.8]} \\
\new{All 14} & \new{absent minus present} & \new{text first} & \new{44{,}786} & \new{report effect} & \new{$+7.8$ [6.9, 8.8]} & \new{2.6 [2.4, 2.9]} \\
\midrule
One & present & image first & 3{,}199 & 1.41\% [0.99, 1.90] & reference & reference \\
One & absent  & image first & 3{,}199 & 10.28\% [8.72, 12.02] & $+8.9$ [7.5, 10.6] & 7.3 [5.3, 10.5] \\
One & present & text first  & 3{,}199 & 2.84\% [2.12, 3.71] & $+1.4$ [0.6, 2.3] & 2.0 [1.4, 3.1] \\
\bottomrule
\end{tabular}}
\caption{Primary MIMIC-CXR comparison and secondary prompt-order analysis (MedGemma-27B, 3{,}199
cases, 293 patients). Rows within a block share cases, substitutes and template; paired differences
and ratios are against the block's reference row, and the two blocks are different units.
\new{The one-question report-present cell is 1.41\% here, 45 of 3{,}199 cases, and 1.38\%, 44
cases, in Tables~\ref{tab:readoutmain}, \ref{tab:textctl} and \ref{tab:readout}, which rerun that
arm; the two runs differ on one case. The two all-14 text-first rate rows are paired against the
all-14 image-first row of the same report condition, on identical case ids; the third gives the
report effect within text first, so the two orders' report effects can be read side by side.} Brackets:
95\% patient-clustered percentile bootstrap, \new{10{,}000 draws for
the all-14 rows and 2{,}000 for the rest}.}
\label{tab:designs}
\end{table}

\textbf{Sampling the discordant image.} The substitute is chosen once per case: uniformly among the
patient's other frontal test-split studies whose CheXpert positive set differs from the source,
failing that any other study of the patient, and for the 19 single-study patients a differently
labelled image from another patient. View and acquisition time are not matched; no pair exceeded
the specified SSIM threshold of 0.9 (Appendix~\ref{app:distance} and Table~\ref{tab:pairing}).
Trials on which the
substitution changes the asked finding's CheXpert label are \emph{label-changed} and reported
separately. That label is not visual ground truth: CheXpert labels come from report
text~\citep{irvin2019chexpert}, which the model reads in the concordant condition, and the
substitute's label comes from a report it never sees. Uncertain and absent labels count as
negative; relabelling cannot change the flip rate, and under alternative policies the label-changed
against label-unchanged difference never separates from zero (Appendix~\ref{app:cohort}).

\textbf{Prompt and readout.} The model's own chat template, one user turn, no system prompt: an
image block, then a text block reading \texttt{Report: <report>}, a blank line, and
\texttt{Question: <question>}; text first swaps the two blocks. Nothing tells the model the report
is historical \new{or that it may be unreliable}; \new{instructions that treat} it as historical
context \new{and that warn it may be wrong are} tested in
Appendix~\ref{app:histframe}. The question is a fixed template per finding with no answer-format
instruction (Appendix Table~\ref{tab:questions}). The \emph{lowercase first-token readout} is the
argmax over the logits of the lowercase \texttt{yes} and \texttt{no} tokens at the first generated
position under greedy decoding; its change under substitution is the \emph{flip rate}. Those tokens
(ids in Appendix~\ref{app:readout}) carry negligible probability mass, which sits on the
capitalised variants, so the readout is a proxy. A validation under an explicit answer instruction
compares it, \new{on both designs,} with the token-family and generated-answer readouts under an
explicit instruction to answer Yes or No (Section~\ref{sec:res1}, Appendix~\ref{app:readout});
\new{the headline prompt carries no such instruction.} In the ablation arms of Appendix Table~\ref{tab:ablate}, deleting the image builds no
image block, so none
reaches the model, and deleting the report removes the ``Report:'' line; nothing else changes. One
image per case, so a report-level label may refer to a view the model was not shown.

\textbf{Statistics.} Intervals for the MIMIC-CXR analyses on the full cohort are patient-clustered
percentile bootstraps over the 293 patients: each bootstrap sample retains all trials belonging to
each sampled patient, and paired differences and ratios are computed inside each draw; \new{10{,}000
draws for the primary all-14 rates and for Appendix Table~\ref{tab:ablate}, 2{,}000 elsewhere.} Subsets and other datasets state
their resampling unit and method where they appear. The label-changed against label-unchanged
comparison
is a two-sided test that can reject equality but not establish it; no equivalence margin was
pre-specified. Per-finding, per-category and per-layer analyses are exploratory and uncorrected for
multiplicity.

\section{Experimental setup}
MedGemma-27B, the instruction-tuned release at revision 2d3e00e~\citep{medgemma2025}, built on
Gemma 3~\citep{gemma3}: 62 decoder layers, residual width 5376, bfloat16, one NVIDIA H200 or H100
per run \new{under Python 3.11, PyTorch 2.10 and Transformers 5.3, with the exact versions of every
package in each run record's environment file. Experimental chronology, for reproducibility: the
lowercase first-token readout under the plain prompt was run first; the explicit answer
instruction, generated-answer scoring, the additional model families and the order rerun
followed, and each is dated in its run record}; images enter at the processor's default preprocessing (896 by 896 pixels, 256 tokens,
pan-and-scan off). MIMIC-CXR~\citep{johnson2019mimiccxr} is the primary dataset; its free-text
reports pair with the 14 CheXpert labels~\citep{irvin2019chexpert} the questions are built from.
VQA-RAD~\citep{lau2018vqarad}, SLAKE~\citep{liu2021slake}, OmniMedVQA~\citep{hu2024omnimedvqa} and
ProbMed~\citep{yan2025probmed} have no report (Appendix Table~\ref{tab:data}). MIMIC-CXR was used
under the PhysioNet Credentialed Health Data License 1.5.0 and its data use agreement; VQA-RAD is
the CC0 1.0 release archived on OSF, SLAKE the authors' CC BY 4.0 release, OmniMedVQA its public
open-access release under its published terms, which pass through each source dataset's licence,
and ProbMed its gated release under its dataset-card terms, an MIT licence over images that keep
their upstream licences. MedGemma weights were used under the Health AI Developer Foundations
terms of use, \new{Qwen3.5-9B and Qwen3.5-27B~\citep{qwen2026qwen35}, LLaVA-NeXT on
Mistral-7B~\citep{liu2024llavanext,liu2023llava}} and the TorchXRayVision
classifier under their Apache 2.0 licences. No
dataset was redistributed, and no image or report left institutional hardware.

\section{Results}\label{sec:results}
\subsection{Report availability and image sensitivity}\label{sec:res1}
\textbf{The report effect.} \new{Under the explicit answer instruction, the model's generated
answer changes with the image on 20.94\% of trials without the report against 4.26\% with it, a
paired difference of 16.7 points [15.6, 17.7] (Table~\ref{tab:readout14}); this is the primary
estimate, and the token-family and lowercase readouts under the same instruction agree to within
0.6 points. The original prompt with the lowercase first-token readout, the paper's first
measurement, gives 17.07\% against 4.70\%, 12.37 points [11.19, 13.57] and a ratio of 3.6
(Table~\ref{tab:designs}; Figure~\ref{fig:abstract}C), and is kept as a sensitivity analysis:
adding the instruction changes the prompt, so the instructed estimate establishes the effect under
that protocol rather than validating every output of the original one.} The one-question design
agrees in direction. \new{The composite question's wording, whether the
image shows any acute cardiopulmonary finding, does not match its target, which is \texttt{yes}
whenever any of the 13 finding-specific targets is, support devices and fracture included.
Dropping it leaves the 13 finding-specific questions, 41{,}587 trials, and gives 4.78\% with the
report against 18.14\% without, a paired difference of 13.36 points [12.16, 14.57] under the
lowercase readout, and 4.42\% [4.08, 4.77] against 19.93\% [18.79, 21.07], 15.5 points [14.5,
16.5], under the generated answer, so the mismatch does not carry the effect and the 13
finding-specific questions are the analysis of record, with the 14-question figures kept for
comparison (Appendix~\ref{app:cohort}).} \new{The report-absent rate is
not uniform over findings, and the pooled 17.07\% hides which ones carry it: five of the 14 sit at
or below 4.1\% because without the report the model answers yes on nearly every trial, at a yes
rate of 97.8\% to 100\% (atelectasis and support devices flip on no trial at all), while the other
nine run from 16.8\% to 32.0\%. The five are the same findings whose image-only specificity is at
or near zero in Appendix Table~\ref{tab:sensspec14}, so the pooled effect varies substantially across findings: several report-absent readouts are
nearly constant, and atelectasis and support devices show lower flip rates without the report
than with it. The pooled increase should not be read as a uniform finding-level effect; the same pattern holds under the generated answer (Appendix Table~\ref{tab:genperfinding}), where atelectasis is 6.81\% with the report against 4.63\% without and support devices 11.03\% against 1.44\%.} On VQA-RAD \new{($n=52$)}, which has no
report and \new{questions the key marks} image-decisive, the flip rate under a final-token yes/no
readout is 61.5\% \new{[48.1, 75.0]} with
accuracy 0.731 on both images.

\begin{table}[H]
\begin{newpar}[breakable=false]
\centering\small
\setlength{\tabcolsep}{4pt}
\adjustbox{max width=\textwidth}{%
\begin{tabular}{lrrrr}
\toprule
Readout & Report present & Report absent & Paired difference, points & Ratio \\
\midrule
Lowercase first token & 4.64\% [4.29, 5.01] & 20.72\% [19.46, 21.93] & $+16.1$ [14.9, 17.2] & 4.5 \\
Token families & 4.30\% [3.99, 4.64] & 21.03\% [19.79, 22.27] & $+16.7$ [15.7, 17.8] & 4.9 \\
Generated answer & 4.26\% [3.95, 4.60] & 20.94\% [19.70, 22.13] & $+16.7$ [15.6, 17.7] & 4.9 \\
\midrule
Lowercase first token, headline prompt & 4.70\% [4.35, 5.06] & 17.07\% [15.78, 18.39] & $+12.4$ [11.2, 13.6] & 3.6 \\
\bottomrule
\end{tabular}}
\caption{All-14 design under the instruction to answer Yes or No (44{,}786 trials, 3{,}199 cases,
293 patients): flip rate under three readouts, paired difference absent minus present in points,
ratio without interval; last row, the headline prompt without the instruction. Brackets: 95\%
patient-clustered percentile bootstrap, \new{10{,}000 draws for the
lowercase headline row and 2{,}000 for the rest}.}
\label{tab:readout14}
\end{newpar}
\end{table}

\begin{table}[!htb]
\centering\small
\setlength{\tabcolsep}{4pt}
\adjustbox{max width=\textwidth}{%
\begin{tabular}{lcc}
\toprule
 & MedGemma-27B & MedGemma-4B \\
\midrule
Concordant accuracy & 0.648 [0.635, 0.662] & 0.673 [0.661, 0.686] \\
Balanced accuracy & 0.717 [0.710, 0.725] & 0.708 [0.701, 0.714] \\
Discordant accuracy & 0.589 [0.576, 0.602] & 0.610 [0.598, 0.624] \\
Flip rate, label changed & 4.30\% [3.78, 4.85] & 7.28\% [6.59, 7.98] \\
Flip rate, label unchanged & 4.81\% [4.45, 5.19] & 7.46\% [6.93, 7.97] \\
Flip rate, all trials & 4.70\% [4.35, 5.06] & 7.42\% [6.94, 7.88] \\
\bottomrule
\end{tabular}}
\caption{All 14 questions, report present, image first, MIMIC-CXR test (44{,}786 trials, 3{,}199
cases, 293 patients). Each image is scored against its own study's report-derived label for the
asked question; balanced accuracy is the mean of the yes and no recalls (majority 0.796). Brackets:
95\% patient-clustered percentile bootstrap, \new{10{,}000 draws}.}
\label{tab:head}
\end{table}

\begin{figure}[!htb]
\centering
\includegraphics[width=\textwidth]{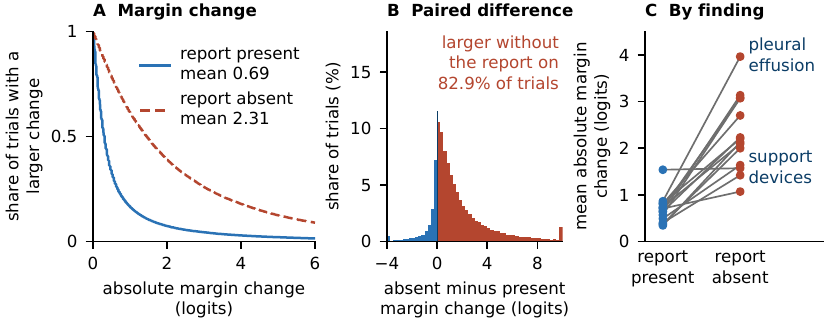}
\caption{\textbf{(A)} Share of the 44{,}786 trials whose absolute margin change $|m_d-m_c|$
exceeds $x$, with and without the report. \textbf{(B)} Paired difference on the same trial, report
absent minus report present; tails fold into the end bins. \textbf{(C)} Mean absolute margin change
per finding, report present to report absent.}
\label{fig:main}
\end{figure}

\textbf{Continuous margins.} We measure changes in the continuous margin
$\log p(\texttt{yes})-\log p(\texttt{no})$ across all trials, including those without a prediction
flip. With the report present the mean absolute paired
change is 0.690 [0.657, 0.726]; 32.2\% of trials move by more than 0.5, against 4.7\% that flip,
and so do 33.0\% of label-unchanged trials. Without the report the mean change is 2.307, larger
than with it on 82.9\% of trials (Figure~\ref{fig:main}A,B). The rise holds on every finding except
support devices, whose mean change is 1.54 with the report and 1.57 without; pleural effusion moves
most, 0.86 against 3.96 (Figure~\ref{fig:main}C). On label-changed trials the
target-aligned change $y_d(m_d-m_c)$, with $y_d\in\{-1,+1\}$ the substitute's label, is 0.35 with
the report and 1.44 without, positive on 57.9\% and 65.4\% of trials (Appendix~\ref{app:cohort},
Figure~\ref{fig:aligned}).

\textbf{Label alignment and transitions.} With the report present the flip rate is about the same
on label-changed and label-unchanged trials (Table~\ref{tab:head}), a clustered difference of
$-0.50$ points [$-1.03$, $+0.02$]; no equivalence is claimed. Scored against each study's own
report-derived label for the asked finding, accuracy is 0.648 on the original image and 0.589 on
the substitute [0.576, 0.602] (Table~\ref{tab:head}); 61.3\% of label-changed trials go from
correct on the original image to wrong on the substitute and 34.4\% the reverse (Appendix
Table~\ref{tab:transition}).

\textbf{Robustness.} The 4B model flips 2.7 points more (Table~\ref{tab:head}), \new{and under the
answer instruction the paired report-presence difference is 13.3 points [11.2, 15.6] in Qwen3.5-9B,
19.60 points [17.00, 22.40] in Qwen3.5-27B and 12.75 points [11.10, 14.44] in LLaVA-NeXT on
Mistral-7B, a third lineage, with image reading on that question set at chance without the report
in every one of them (Appendix~\ref{app:family}).} \new{Redrawing the substitute image under two further seeds changes
the one-question report effect by at most 0.34 points (Appendix~\ref{app:cohort}).} The flip rate rises
with the embedding distance between the images in both conditions; adjusting for distance and label
change, the odds ratio for report presence is 0.12 [0.09, 0.17] (Appendix~\ref{app:distance}). Where the classifier surrogate predicts a change in the queried
finding, the report-absent flip rate is three times that where it predicts the same label for both
images (Appendix~\ref{app:clf}).
Under manually annotated report labels the effect persists: on the 800 trials whose manual label
changes, agreement with the substitute's label is 0.378 with the report and 0.519 without
(Appendix~\ref{app:rad}). It holds under a second question template, with a template-by-report
interaction of 1.7 points [$-0.0$, 3.4] (Appendix~\ref{app:template}). It holds with the answer
options in either order, with an order-by-report interaction of $-0.53$ points [$-1.36$, 0.30]
(Appendix~\ref{app:ansorder}), and with an explicit Uncertain option offered
(Appendix~\ref{app:abstain}). \new{Warning the model that the report may not describe the image and
may be wrong leaves the effect intact: the one-question flip rate is 1.50\% under the warning
against 1.22\% under the plain framing with the same answer instruction, a paired difference of
$+0.28$ points [$-0.13$, 0.74] (Appendix~\ref{app:histframe}).} \new{On the headline unit, text-block order does not move the report-present rate, $+0.2$
points [$-0.1$, 0.5], and lowers the report-absent rate by 4.3 points; on the one-question design
the report-present rate doubles (Table~\ref{tab:designs}; Appendix~\ref{app:textctl}).}

\textbf{Readout validation.} This validation uses an explicit answer instruction and covers
\new{both designs; the headline estimate} of 4.70\% against 17.07\% is a result for the lowercase
first-token readout \new{under a prompt that carries no answer instruction.} \new{On the
one-question design, with the}
instruction to answer Yes or No appended, the lowercase readout agrees with the generated answer on
98.8\% of report-present and 91 to 93\% of report-absent cases, and the three readouts agree on the
report-presence difference (Table~\ref{tab:readoutmain}). Without the instruction the lowercase
readout gives a smaller report-absent rate than the token-family readout, so the report-absent
value depends on the readout choice; the direction of the report effect does not
(Appendix~\ref{app:readout}). \new{The primary effect survives generated-answer scoring on its own
unit: rerun with the same instruction, the all-14 design gives a generated-answer flip rate of
4.26\% with the report against 20.94\% without, a paired difference of 16.7 points [15.6, 17.7]
(Appendix Table~\ref{tab:readout14}). The instruction raises the report-absent rate, from 17.07\% to
20.72\% under the lowercase readout on the same trials, and leaves the report-present rate within a
tenth of a point of 4.70\%, so the instructed prompt is not the headline prompt and the headline
numbers stand. Under every readout the direction is the same and every interval on the difference
excludes zero, with ratios of 4.5 to 4.9.}

\begin{table}[!htb]
\centering\small
\setlength{\tabcolsep}{4pt}
\adjustbox{max width=\textwidth}{%
\begin{tabular}{llrrr}
\toprule
Prompt & Readout & Report present & Report absent & Paired difference, points \\
\midrule
With instruction & lowercase first token & 1.22\% & 18.26\% & $-17.0$ [$-19.4$, $-14.9$] \\
With instruction & token families & 0.84\% & 17.69\% & $-16.9$ [$-18.8$, $-15.0$] \\
With instruction & generated answer & 0.84\% & 17.66\% & $-16.8$ [$-18.7$, $-15.0$] \\
No instruction & lowercase first token & 1.38\% & 10.28\% & $-8.9$ [$-10.6$, $-7.5$] \\
No instruction & token families & 1.09\% & 17.35\% & $-16.3$ [$-18.5$, $-14.2$] \\
No instruction & generated answer & 1.84\% & 11.57\% & $-9.7$ [$-11.5$, $-8.1$] \\
\bottomrule
\end{tabular}}
\caption{Readout validation, secondary one-question design (3{,}199 cases, 293 patients): flip
rate under three readouts and two prompts, present minus absent in points. Per-cell intervals in
Appendix Table~\ref{tab:readout}; \new{the all-14 design under the instruction is in Appendix
Table~\ref{tab:readout14}.} Brackets: 95\% patient-clustered percentile bootstrap, 2{,}000 draws.}
\label{tab:readoutmain}
\end{table}

\textbf{Text content.} Text content also affects image sensitivity. In the one-question design,
length-matched neutral prose and clinical boilerplate produce flip rates of 8.2\% and 7.1\%,
compared with 10.3\% without a report and 1.4\% with the full report. Removing the sentences that
mention the queried finding raises the rate to 11.6\%. A sentence stating the reference answer
reduces it to 3.1\%, and replacing the target sentences with the opposite label reduces
source-label accuracy to 4.3\% (Appendix Table~\ref{tab:textctl}, Figure~\ref{fig:textpos}). These
controls suggest that answer-bearing text contributes to the reduced sensitivity, although they do
not isolate its contribution from changes in length and context.

\subsection{Modality ablation: report versus image}\label{sec:res2}
On the all-14 design, scored against report-derived labels, the report alone reaches a pooled
balanced accuracy of 0.770; adding the image lowers it to 0.717, the image with the question gives
0.665, and the question alone is at chance, answering yes throughout (Appendix
Table~\ref{tab:cells14}). The two inputs interact: pooled over trials, the image raises balanced
accuracy by 0.165 [0.158, 0.172] without the report and lowers it by 0.053 with it, a pooled
interaction of $-0.218$. Macro-averaged over the 14 findings the image adds 0.076 without the
report and the interaction is $-0.128$ [$-0.141$, $-0.115$], still negative. Pooling over trials
matches the paper's estimand, the average over trials; the macro average weights each finding
equally and shows that the image-only gain is not uniform, since several findings sit at chance in
that cell. Report only is the best cell on 9 of the 14 findings. The
one-question design (Appendix Table~\ref{tab:ablate}) also ranks report only above the full arm.
\new{Both designs are scored against report-derived labels, so a fall when the image is added
measures movement away from the report's label and not a visual error; neither the all-14
interaction nor the one-question ranking shows that the image misleads.}

\subsection{Secondary analyses}\label{sec:res3}
\textbf{Per-finding agreement, chance-corrected.} Raw agreement rewards a skewed responder, so we
also report Cohen's $\kappa$~\citep{cohen1960kappa} per finding (Appendix Table~\ref{tab:floor};
all 14 in Table~\ref{tab:sup-kappa}). Against chance the ranking changes: for support devices,
agreement is 0.946 and $\kappa$ is 0.370, less than half the next lowest, because predictions are
strongly skewed toward yes.

\textbf{Flip rate by question type.} On SLAKE, with no report, perceptual questions flip more often
than knowledge questions: organ 81.0\% and modality 80.0\% against abnormality 36.0\% and
knowledge-graph 25.9\% (Appendix Table~\ref{tab:slake}). Sample sizes vary by category and no
ordered-alternative test was run; comparisons are exploratory.

\subsection{\new{Image sensitivity across decoder layers}}\label{sec:layers}
\begin{newpar}
We next examined how image substitutions affect projected answer scores across decoder
layers, with and without the report. We evaluated these trajectories alongside agreement with the
final answer to assess where the intermediate readout was informative. We then used attention
interventions to test the contribution of direct access to report-token positions
(Figure~\ref{fig:layers}; full analysis in Appendix~\ref{app:neg}).

At each layer we project the residual stream through the output head and take the yes-minus-no
margin, then average the absolute change in that margin between the two images over the 44{,}786
trials (Figure~\ref{fig:layers}A). The change is small through the middle of the network and
rises from about layer 46: with the report it is 0.04 at layer 25 and 1.31
at layer 46, ending at the published 0.690 at the output; without the report it is 0.05 and
4.54, ending at 2.307. The report-absent trajectory sits above the report-present one at
every layer from 46 onward, which is the layerwise form of the headline effect. Figure~\ref{fig:layers}B
shows how far each layer's projected answer agrees with the final answer: agreement is 0.55
with the report and 0.67 without at layer 25, against a majority-class floor, and it
reaches 0.95 and 0.93 at layer 46. The intermediate readout carries a
limited, non-persistent signal before layer 46 and tracks the final answer after it, so the
layer-25 feature reported in Appendix~\ref{app:neg} is retained as an observation about the
projection and not read as the point at which the answer is decided. Figure~\ref{fig:layers}C
shows the attention intervention against its two controls; the trajectory and the intervention
answer different questions, one describing projected responses across depth and the other the
effect of interrupting a specified pathway, and Section~\ref{sec:depth} states what the
intervention does and does not establish.
\end{newpar}

\begin{figure}[t]
\centering
\includegraphics[width=\textwidth]{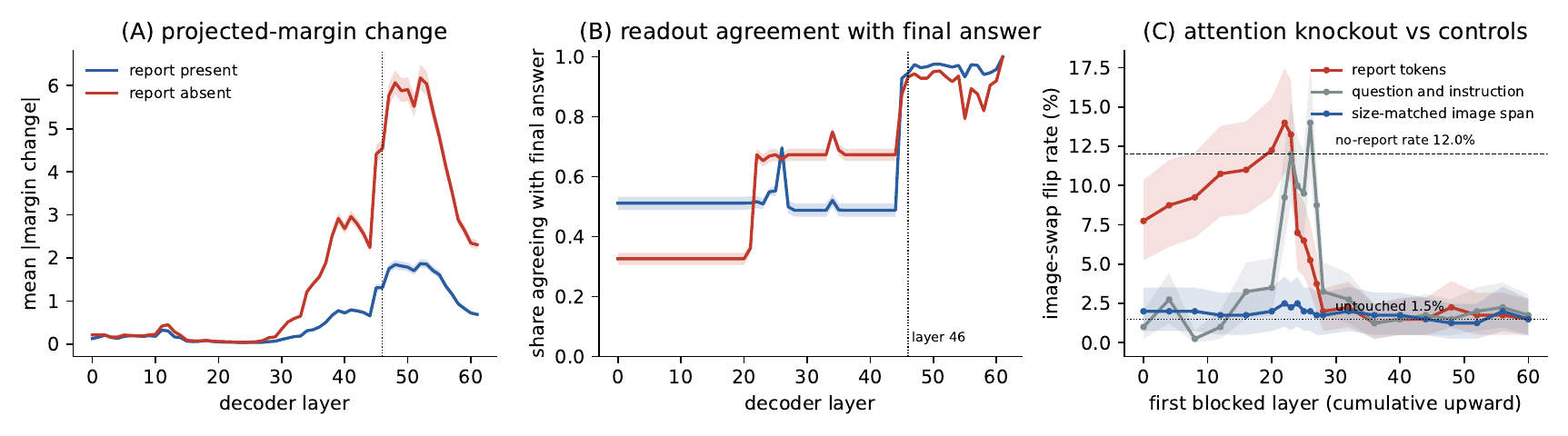}
\begin{newpar}[breakable=false]
\caption{\textbf{(A)} Mean absolute change in the projected yes/no margin between the two images
at each decoder layer, report present and absent, 44{,}786 trials, patient-clustered 95\%
intervals; the layer-61 values are the published 0.690 and 2.307. \textbf{(B)} Share of trials
whose projected answer at each layer equals the final answer. \textbf{(C)} Report-token attention
knockout, cumulative from the plotted layer upward, against the question-and-instruction span and
a size-matched image span, 400 one-question cases.}
\label{fig:layers}
\end{newpar}
\end{figure}

\subsection{\new{At what depth does the report take hold?}}\label{sec:depth}
\new{Cumulative blocking of attention to report-token positions increased image-swap sensitivity
when initiated at early decoder layers, and the effect diminished for interventions initiated
later. These results identify when direct access to report-token positions affects the measured
response; they do not localise all downstream use of report information, since report-derived
content may already have been copied into other positions before a later block starts. The
estimates: masking the text stream's attention to the report raises the image-swap flip rate from
1.50\% to 7.75\% when the block begins at layer 0, against 12.00\% with no report, and is at
baseline from layer 32. Single-layer blocking never exceeds 2.50\%, which is consistent with
redundant or distributed pathways and does not show that no layer carries the effect. The 14.00\%
at layer 22 overshoots the no-report rate and falls in the window where the question-and-instruction
control also moves, so the report-specific comparison is confined to layers 0 to 20, where the
paired difference clears zero against both controls and the two arms flip disjoint case sets. A
higher flip rate under the block is not by itself a restored reading of the substituted image.
Cross-condition residual patching adds no localisation,
being a null against its wrong-donor control, 14.07\% against a correct-donor maximum of 12.25\%
(Appendix~\ref{app:neg}, Figure~\ref{fig:knockout}). This is a MIMIC-CXR result: the image-token knockout run on three
further datasets does not separate from its control on VQA-RAD or OmniMedVQA, and the control
comparison was unavailable for SLAKE, so it is reported as a negative and not generalised.}

\subsection{Negative results: steering and probe transfer}\label{sec:neg}
Two mechanistic analyses were tried and both are negative (Appendix~\ref{app:neg}). Difference-of-means
steering did not move the share of answers that track the substituted image below the scale that
degraded the outputs, where the targeted direction did not separate from a random one, and changed
it by 0.1 points at the largest scale. A linear
probe fit under a frozen protocol and run once on a disjoint patient cohort fell from its test-split
discrimination to chance. The paper makes no \new{localisation} claim\new{ beyond that bound}.

\section{Limitations}\label{sec:limits}
\textbf{Label validity.} The labels are report-derived throughout. The label-changed split
measures consistency with the report, not with the image; the manually annotated report labels of
Appendix~\ref{app:rad} check the automated extraction, not the image. We did not obtain independent
image annotations for this evaluation; the only image-level labels are the classifier surrogate of
Appendix~\ref{app:clf}. The audit says how much the readout moves with the image, and the
opposite-label control says the readout follows the text when the two conflict. Neither says
whether following the text was wrong: a report is an expert reading, and deferring to an accurate
one may be rational. No chronology was supplied to the model, so Appendix~\ref{app:stale} is a
retrospective stratification, in which later and earlier substitutes give nearly the same
report-agreement rate. An instruction asking the model to treat the report as historical context
raised the flip rate by about half a point\new{, and an explicit warning that the report may not
describe the image and may be wrong changed it by $+0.28$ points [$-0.13$, 0.74] against the plain
framing}; \new{these were framing interventions} with no temporal
consistency enforced, so \new{they do} not test the handling of reports that are in fact stale\new{,
and neither restores image sensitivity: under the warning the readout still matches the substituted
image's label on only 8.49\% of the label-changed pairs}
(Appendix~\ref{app:histframe}).

\textbf{Readout validity.} The output is the lowercase first-token readout of a prompt that does
not request a yes or no. It cannot flag conflicting or missing evidence, it records a shift that
does not cross the boundary as no change, and \new{its validation against generated answers, added
after the initial analysis, covers both designs only} under an explicit answer-format instruction
\new{that the headline prompt does not carry} (Appendix~\ref{app:readout}); \new{under that
instruction the primary effect keeps its direction and grows.} The report-absent one-question
rate depends on the readout (10.3\% on the paper's tokens, 17.4\% on the token families); the
report-present rate does not.

\textbf{Sampling and prompt dependence.} One substituted image per case, drawn without matching on
view, positioning or the other thirteen labels, so a substitute can carry the same finding, which
lowers the flip rate even for an image-conditioned model. \new{The flip rate depends on text-block order, and the headline
unit was rerun under both orders to bound that dependence (Table~\ref{tab:designs}). The report
effect survives the swap but shrinks: $+7.8$ points [6.9, 8.8] with the text block first against
$+12.4$ [11.2, 13.6] with the image first, on identical trials, so the direction and its
significance are robust to order while the magnitude is not. The order effect itself is a null
with the report present, $+0.2$ points [$-0.1$, 0.5], which does not reproduce the
one-question design's $+1.4$ [0.6, 2.3], and it is negative with the report absent, $-4.3$ points
[$-5.4$, $-3.3$]. That pooled figure is a net of large offsetting per-finding effects rather than a
uniform shift: text first lowers the report-absent rate on nine of the 14 findings and raises it on
five, over a range from $-21.7$ to $+14.3$ points, and support devices moves from flipping on no
trial to 8.07\%. No flip rate in this paper should be read as an order-independent property of the
model.} \new{Every report-availability estimate here comes from one
dataset, the MIMIC-CXR test split, 293 patients at a single institution, frontal chest radiographs
and English reports. No other dataset used in this paper pairs an image with a report, so the
manipulated factor is untested outside that setting.}

\textbf{Generalisation.} \new{The headline rests on} one model family and one snapshot; the model
card lists MIMIC-CXR and SLAKE in training, which does not by itself establish exposure to these
test cases. MedGemma-4B shows the same interaction with a larger effect. \new{The effect is not
specific to one lineage or one size: under the answer instruction it replicates in direction and
significance across three families, MedGemma, Qwen3.5 at 9B and 27B, and LLaVA-NeXT on Mistral-7B.
On the label-conditioned one-question set no model tested reads the substituted image above
chance once the report is removed, so what generalises there is report anchoring, not image
competence; on the all-14 unit the image does carry signal for MedGemma, which is why that claim
is stated for the one-question set only (Section~\ref{sec:res2}, Appendix~\ref{app:family}).}
Every layerwise readout uses the
uncalibrated logit lens: the tuned lens~\citep{belrose2023tunedlens} scored below it on both
readout datasets (Appendix Table~\ref{tab:sup-lenseval}), and activation
patching~\citep{meng2022rome} leaves the depth under-determined, so no depth claim \new{rests on a
projection, and the knockout below identifies when direct access to report-token positions
matters rather than where report information stops being used}.
\new{A calibration pass puts the shallowest depth at which the projection tracks the model's own
answer at layer 46 of 62 (Appendix~\ref{app:neg}); everything shallower, the layer-25 feature of
Figure~\ref{fig:layer} included, sits where the projection does not clear the best constant
predictor of that answer, and that feature is a
property of the report-present prompt rather than of the network. The one depth statement the
paper does make is causal rather than projected, and it is a bound and not a locus: a cumulative
attention block on the report restores image sensitivity when it begins at or below layer 22 and
not from layer 28 upward, report-specific over layers 0 to 20, so the boundary sits near layer 24
of 62; the knockout is not equivalent to deleting the report, no single layer carries the effect,
and cross-condition residual patching does not separate from a wrong-donor control at any depth
(Appendix~\ref{app:neg}).} These
metrics say \emph{whether} the readout changes with the image, not competence.

\section{Conclusion}
\new{On 44{,}786 paired MIMIC-CXR trials, removing the report raises the rate at which
MedGemma-27B's generated answer changes with the image from 4.26\% to 20.94\%, a paired
difference of 16.7 points [15.6, 17.7] under an explicit answer instruction; the original prompt
with a lowercase first-token readout gives 4.70\% against 17.07\%, 12.37 points [11.19,
13.57].} Continuous answer margins are also more sensitive to the image without the report.
The direction of the report effect persists across the evaluated controls, \new{under
generated-answer scoring of both designs, and across three model families and two model sizes
within \new{two} of them; since none of those models shows above-chance reading of the substituted image on the
one-question set\new{; on six finding-specific questions asked of every case, chosen by a prespecified prevalence rule so the selection cannot score on the label, the report effect holds at 13.1 to 20.9 points in all three (Appendix Table~\ref{tab:families13q})}, what replicates there is report anchoring rather than image competence.} These findings show that report
availability changes measured image sensitivity; independent image annotations are needed to
determine when this behaviour produces visually incorrect answers.

\typeout{MODALENS body ends on page \thepage}

\clearpage
\raggedbottom

\section*{Acknowledgements}
This research was supported by Anthropic's AI for Science program, by GPUs from NVIDIA Brev,
and by the MIT ORCD cluster.

\bibliographystyle{plainnat}
\bibliography{references}

\appendix
\section{Datasets, cohort and questions}
\begin{table}[H]
\centering\small
\setlength{\tabcolsep}{4pt}
\adjustbox{max width=\textwidth}{%
\begin{tabular}{lll}
\toprule
Dataset & Role in this study & Additional clinical context \\
\midrule
MIMIC-CXR~\citep{johnson2019mimiccxr}   & primary dataset, 14 questions per case & free-text report \\
VQA-RAD~\citep{lau2018vqarad}           & image-decisive positive control & none ($n{=}52$) \\
SLAKE~\citep{liu2021slake}              & question-type breakdown & none \\
OmniMedVQA~\citep{hu2024omnimedvqa}     & modality breadth, 4-way MCQ & none \\
ProbMed~\citep{yan2025probmed}          & false-premise question axis & none (false-premise question) \\
\bottomrule
\end{tabular}}
\caption{\new{Datasets used. MIMIC-CXR, the only one pairing each image with a free-text report,
carries the report-availability comparison and every headline number; VQA-RAD and SLAKE also
appear in the body, as an image-decisive control and a question-type breakdown
(Sections~\ref{sec:res1} and \ref{sec:res3}); OmniMedVQA and ProbMed appear only in the
supplementary tables.}}
\label{tab:data}
\end{table}

\begin{table}[H]
\centering\small
\setlength{\tabcolsep}{4pt}
\adjustbox{max width=\textwidth}{%
\begin{tabular}{llrr}
\toprule
Group & Item & & \\
\midrule
Swap tier & Same patient, different study & 3{,}180 & 99.4\% \\
 & Different patient, different label & 19 & 0.6\% \\
\midrule
Time gap, source to substitute study (days) & Minimum / Q1 & 0.007 & 15.0 \\
 & Median / mean & 141.3 & 342.0 \\
 & Q3 / maximum & 542.1 & 1{,}829 \\
 & Substitute earlier / later & 1{,}612 & 1{,}568 \\
 & Gap under 1 day / under 7 days & 3.2\% & 17.9\% \\
 & Gap under 30 days / over 365 days & 32.8\% & 33.7\% \\
\midrule
View position, source $\rightarrow$ substitute & AP $\rightarrow$ AP / AP $\rightarrow$ PA & 1{,}546 & 406 \\
 & PA $\rightarrow$ PA / PA $\rightarrow$ AP & 363 & 431 \\
 & Either view unlabelled & 453 & \\
 & Same view, both known & 1{,}909 / 2{,}746 & 69.5\% \\
\midrule
CheXpert labels differing (of 14) & None / exactly one & 22 & 402 \\
 & Exactly two / three or more & 937 & 1{,}838 \\
 & Median / mean & 3 & 2.89 \\
 & Asked answer changes (own finding) & 2{,}321 & 72.6\% \\
\midrule
Reuse and reciprocity & Distinct substitute images / studies & 2{,}093 & 1{,}953 \\
 & Maximum reuse of one image / study & 6 & 8 \\
 & Pairs whose reverse pair also exists & 376 & 11.8\% \\
 & Source studies / with more than one pair & 2{,}816 & 344 \\
\bottomrule
\end{tabular}}
\caption{Composition of the 3{,}199 MIMIC-CXR discordant pairs (seed 0). Time gaps use the 3{,}180
same-patient pairs. Label differences count CheXpert findings whose positive status differs between
source and substitute.}
\label{tab:pairing}
\end{table}

\begin{table}[H]
\centering\small
\setlength{\tabcolsep}{4pt}
\adjustbox{max width=\textwidth}{%
\begin{tabular}{>{\raggedright\arraybackslash}p{0.33\textwidth}>{\raggedright\arraybackslash}p{0.30\textwidth}>{\raggedright\arraybackslash}p{0.30\textwidth}}
\toprule
Question & Question text & Target \\
\midrule
13 finding-specific: Atelectasis, Cardiomegaly, Consolidation, Edema, Enlarged Cardiomediastinum, Fracture, Lung Lesion, Lung Opacity, Pleural Effusion, Pleural Other, Pneumonia, Pneumothorax, Support Devices & \emph{Does this chest X-ray show <finding>?}, finding in lower case, e.g. \emph{Does this chest X-ray show support devices?} & \texttt{yes} iff the finding is in the image's CheXpert positive set (report-derived) \\
\midrule
Composite (any finding) & \emph{Does this chest X-ray show any acute cardiopulmonary finding?} & \texttt{yes} iff any of the 13 finding-specific targets is \texttt{yes}, \texttt{no} otherwise; the dataset's No Finding label is not used \\
\bottomrule
\end{tabular}}
\caption{The 14 question templates and target rules, 13 finding-specific and one composite. The
one-question design asks each case only its primary positive finding, the first positive in fixed
CheXpert order, or the composite question with target \texttt{no} on the 492 cases with none. No
answer-format instruction is appended.}
\label{tab:questions}
\end{table}

\begin{table}[H]
\centering\small
\setlength{\tabcolsep}{4pt}
\adjustbox{max width=\textwidth}{%
\begin{tabular}{lccc}
\toprule
Content type & $n$ & Flip rate & Concordant acc. \\
\midrule
Organ           & 42 & 80.95\% [69.05, 90.48] & 0.857 [0.738, 0.952] \\
Modality        & 20 & 80.00\% [60.00, 95.00] & 0.900 [0.750, 1.000] \\
Position        & 10 & 40.00\% [10.00, 70.00] & 0.700 [0.400, 1.000] \\
Abnormality     & 50 & 36.00\% [24.00, 50.00] & 0.640 [0.500, 0.760] \\
Knowledge-graph & 54 & 25.93\% [14.81, 38.89] & 0.593 [0.463, 0.722] \\
\bottomrule
\end{tabular}}
\caption{SLAKE by question type, no report, 176 of the 186 cases; Plane ($n{=}8$) and Color
($n{=}2$) are omitted for size. Sample sizes vary by category; comparisons are exploratory.
\new{Brackets: 95\% percentile bootstrap over the cases in each category, 5{,}000 draws.}}
\label{tab:slake}
\end{table}

\begin{figure}[H]
\centering
\includegraphics[width=0.78\textwidth]{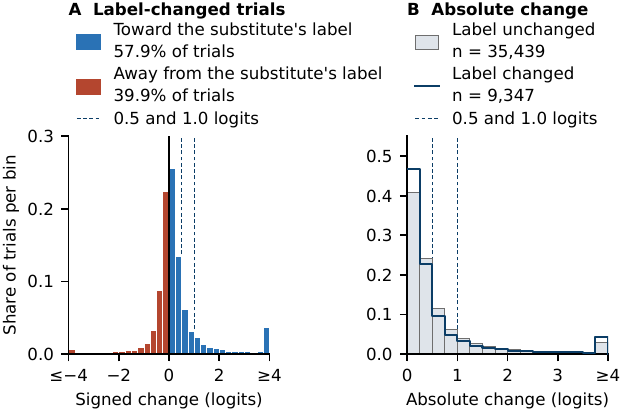}
\caption{Paired margin change, all 14 questions, report present, image first. \textbf{(A)} Signed,
target-aligned change $y_d(m_d-m_c)$ on the 9{,}347 label-changed trials, shaded by direction; mean
0.35 [0.29, 0.41], median 0.08. \textbf{(B)} Absolute change $|m_d-m_c|$ on the 35{,}439
label-unchanged and the same 9{,}347 label-changed trials. End bins collect the tails.}
\label{fig:aligned}
\end{figure}

\begin{figure}[H]
\centering
\includegraphics[width=0.78\textwidth]{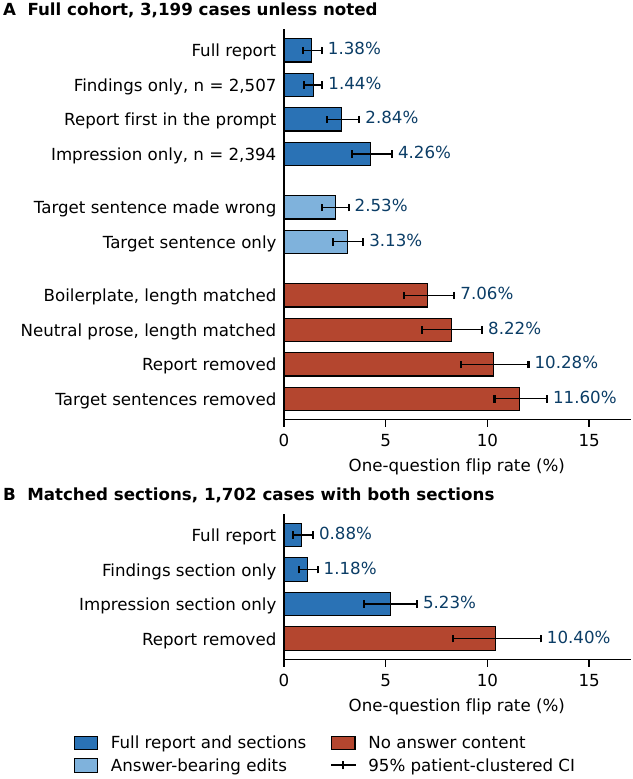}
\caption{One-question flip rate by text condition, MedGemma-27B, patient-clustered 95\% intervals.
\textbf{(A)} Full cohort, 3{,}199 cases unless noted (Appendix Table~\ref{tab:textctl}).
\textbf{(B)} The 1{,}702 cases with both sections, all scored on the same cases:
findings alone within 0.3 points of the full report, impression alone 4.4 points above it [3.0,
5.7].}
\label{fig:textpos}
\end{figure}

\section{How much the image changes}\label{app:distance}
Two distances between the source and substitute images were computed for all 3{,}199 pairs.
Structural similarity (SSIM) at the model's 896 by 896 greyscale input has quartiles 0.462 / 0.528 /
0.586 and maximum 0.78; no pair exceeded the specified SSIM threshold of 0.9 (fraction above it
0.000).
The cosine distance between MedGemma's vision-tower (SigLIP, mean-pooled) embeddings,
cohort-mean-centred, has quartiles 0.395 / 0.563 / 0.745 and a Spearman correlation of 0.16 with
the time gap. With the report absent (one question per case, 329 flips) the flip rate rises with
embedding distance, from 8.5\% [6.3, 11.0] in the most similar quartile. The most distant quartile
gives 14.5\% [11.7, 17.4]. The difference is $+6.0$ points [2.5, 9.5]. With the report present (45
flips) the rise is from 0.6\% to 1.9\%, a difference of $+1.3$ [0.1, 2.5]. A logistic regression of
flip on standardized embedding distance, label change and report presence (6{,}398 pair-condition
rows, cluster-robust) gives an odds ratio of 1.30 per SD of distance [1.16, 1.46]. The odds ratio
is 1.43 for a label change [1.05, 1.96]. It is 0.12 for report presence [0.09, 0.17].

\section{Per-finding agreement}
\begin{table}[H]
\centering\small
\setlength{\tabcolsep}{4pt}
\adjustbox{max width=\textwidth}{%
\begin{tabular}{lcccc}
\toprule
Finding & Observed agreement & Chance-expected & Raw excess & Cohen's $\kappa$ \\
\midrule
Pneumothorax    & 0.992 \new{[0.988, 0.995]} & 0.919 & 0.073 & 0.899 \new{[0.837, 0.943]} \\
Any finding (composite) & 0.964 \new{[0.957, 0.971]} & 0.501 & 0.463 & 0.927 \new{[0.913, 0.941]} \\
Atelectasis     & 0.951 \new{[0.942, 0.960]} & 0.698 & 0.254 & 0.839 \new{[0.807, 0.867]} \\
Cardiomegaly    & 0.936 \new{[0.923, 0.948]} & 0.500 & 0.435 & 0.871 \new{[0.845, 0.896]} \\
Support devices & 0.946 \new{[0.931, 0.959]} & 0.914 & 0.032 & 0.370 \new{[0.271, 0.451]} \\
\bottomrule
\end{tabular}}
\caption{Five of 14 findings, each $n{=}3{,}199$; pooled observed agreement is 0.953.
Chance-expected agreement is $p_c p_d+(1-p_c)(1-p_d)$ from the condition-specific yes-rates, raw
excess is observed minus expected, $\kappa$ is excess over one minus expected. $\kappa$ corrects
for imbalance, not for grounding. \new{Brackets: 95\% patient-clustered percentile bootstrap over
the 293 patients, 2{,}000 draws at seed 0. $\kappa$ depends on both branches' marginals, so each
draw rebuilds them from the resampled cases rather than averaging a per-case value; the three
indicators are resampled jointly so the pairing is preserved. Support devices is the one finding
whose interval is wide and low, [0.271, 0.451], which is what a 0.914 chance level leaves room
for.}}
\label{tab:floor}
\end{table}

\section{Retrospective chronology analysis of stale reports}\label{app:stale}
No chronology was supplied to the model, so this is a retrospective stratification by chronology,
not a test of whether the model reasons about staleness. By construction of the same-patient
pairing, the model reads the source report and sees another study of the same patient, later than
the report on 1{,}038 pairs and earlier on 1{,}130 among the 2{,}168 same-patient pairs (254
patients) whose asked-finding label differs between the two studies. \new{These 2{,}168 are the
same-patient subset of the 2{,}321 pairs whose swap changes the asked finding
(Table~\ref{tab:pairing}) on which the substitute study also carries a CheXpert value for that
finding: of the 3{,}180 same-patient pairs, 171 have no such value on the substitute and 841
carry the same value, leaving 2{,}168, while the 2{,}321 count is taken over all 3{,}199 pairs and
includes the 19 cross-patient ones. Every pair here has a non-zero study-time gap; none was
dropped for a missing one.} Table~\ref{tab:stale} gives
the agreement rates. With the report present the readout follows the report's label at nearly the
same rate for later and earlier substitutes, and the rate does not move with the time gap. The cell
\new{that most resembles the stale-report scenario}, a finding labelled in a later study's report
that the shown report does not mention, has the lowest report-agreement rate\new{; whether this
corresponds to clinical harm is not measured here}: in this pair set it is the composite
any-finding question only, because the one-question design asks about the source study's own
positive finding.
\begin{table}[H]
\centering\small
\setlength{\tabcolsep}{4pt}
\adjustbox{max width=\textwidth}{%
\begin{tabular}{lrr}
\toprule
 & Agrees with & Agrees with \\
Pairs & report label & substitute label \\
\midrule
All 2{,}168, report present & 0.912 [0.897, 0.926] & 0.088 [0.074, 0.103] \\
All 2{,}168, report absent & & 0.321 [0.289, 0.351] \\
Present minus absent & & $-0.232$ [$-0.262$, $-0.202$] \\
Substitute later (1{,}038), report present & 0.899 [0.878, 0.919] & \\
Substitute earlier (1{,}130), report present & 0.924 [0.904, 0.942] & \\
Later minus earlier & $-0.025$ [$-0.054$, $+0.002$] & \\
Odds ratio per log day of time gap & 0.94 [0.87, 1.01] & \\
Later study's finding not in the shown report (188) & 0.824 [0.762, 0.878] & \\
\bottomrule
\end{tabular}}
\caption{Readout agreement by chronology on the 2{,}168 same-patient pairs whose asked-finding
label differs between the two studies, one-question design, report-derived labels. Brackets: 95\%
patient-clustered percentile bootstrap over the 254 patients, 2{,}000 draws.}
\label{tab:stale}
\end{table}

\textbf{Sensitivity to an instruction treating the report as historical \new{or as
unreliable}.}\label{app:histframe}
We tested an instruction that asked the model to treat the report as historical context. This was
a framing intervention; temporal consistency was not enforced. The report-present condition was
rerun with one sentence placed before the report, stating that it describes an earlier study of
the same patient and that the question is to be answered from the current image with the report
as context only, under the same explicit answer instruction as Appendix~\ref{app:readout}. No
study dates were checked: for the original image the report belongs to the study shown, and a
substitute may precede or follow it. Under this framing the readout changed with the image on
1.72\% of pairs [1.25, 2.25], against
1.22\% under the plain ``Report:'' label with the same instruction. Readout accuracy against the
concordant label stays at 92.0\% on the original image and 92.2\% on the substitute. The paired
difference is $+0.50$ points [0.07, 0.94], with the same sign under the generated-answer readout
($+0.60$ points) and the token-family readout. Against the no-report condition (18.26\%) the gap is
16.54 points [14.46, 18.86], so the framing does not restore the sensitivity seen without the
report. Against the report-present cell without the answer instruction (1.41\%) the difference is
$+0.31$ points [$-0.27$, 0.87]. The result measures sensitivity to the instruction, not performance
when the supplied report is in fact historical.

\begin{newpar}
\textbf{An explicit reliability warning.} The historical framing only relocates the report in
time, so the report-present condition was also rerun under the stronger framing, one sentence
before the report telling the model that the report may not describe this image and may be wrong,
that the question is to be answered from the image and that the report is unverified context,
with the same answer instruction; the readout changed with the image on 1.50\% of pairs [1.06,
1.98]. Against the plain ``Report:'' label with the same instruction (1.22\%) the paired
difference is $+0.28$ points [$-0.13$, 0.74], not distinguishable from zero. Against the
historical framing it is $-0.22$ points [$-0.67$, 0.22], also not distinguishable, and the
token-family (1.47\%) and generated-answer (1.38\%) readouts give the same picture. Against the
no-report condition under the same instruction (18.26\%) the gap is $-16.76$ points [$-18.93$,
$-14.74$], so the warning does not restore the sensitivity seen without the report. On the
substituted image the readout still matches the report's label on 93.12\% of pairs and the shown
image's label on 32.89\%, falling to 8.49\% [7.22, 9.93] on the 2{,}321 pairs whose swap changes
the asked finding. Format compliance
rises to 99.94\% and 99.87\% on the two image sides from 99.28\% and 99.50\% under the historical
framing, and the generated-answer flip rate shifts against the plain framing by $+0.53$ points
[0.12, 0.97].
\end{newpar}

\begin{newpar}
\textbf{On the pairs where the framing is true.} The historical instruction was applied to every
pair, but it is a true statement only where the substitute image is a later study than the one the
shown report describes. Chronology is reconstructed from the signed study gap in the pairing record for every
same-patient pair, which splits the 3{,}180 into 1{,}568 where the substitute is a later study,
so the shown report really is historical relative to the image, and 1{,}612 where it is earlier;
there are no equal-date or undated pairs, and the counts reconcile with Table~\ref{tab:stale}.
Under the historical framing the image-swap flip rate is 1.72\% on the later set and 1.74\% on
the earlier set, a difference of $-0.02$ points [$-1.03$, $+1.06$]; scoring the generated answer
gives 1.66\% against 0.99\%, $+0.67$ [$-0.27$, $+1.66$]. Both intervals come from resampling the
276 patients once per draw and scoring both strata on that draw, since a patient can contribute
pairs to both. We did not detect a difference between the strata: the instruction does not
measurably restore image sensitivity on the pairs where what it says is the case. This is the
nearest this data comes to a stale-report condition without independently dated images; no report
was rewritten, and the absence of a detected difference is not evidence that stale reports are
harmless.
\end{newpar}

\section{Sensitivity analysis using manually annotated report labels}\label{app:rad}
We repeated the analysis using the manually annotated report labels supplied with MIMIC-CXR-JPG:
the 687 test-set reports annotated by a single radiologist to evaluate the automated
labelers~\citep{johnson2019mimiccxrjpg}. This checks sensitivity to automated label extraction, not
the findings in the displayed images: the annotator read the report, not the image. Applied at
study level, 291 of the 3{,}199 pairs (90 patients) have manual labels on both studies, 4{,}074 of
the 44{,}786 trials. On those studies the automated and manual report labels agree on 0.943 [0.938,
0.947] (Cohen's $\kappa$ 0.795; automated sensitivity 0.894, precision 0.773 against the manual
label), which estimates agreement between automated and manual labels in this subset.
Table~\ref{tab:rad} gives the
readout on the 800 trials whose manual label of the asked finding differs between the two studies:
the report effect keeps its direction. Scored against the manual rather than the automated label,
concordant accuracy is 0.627 (balanced 0.712) with the report and 0.485 (balanced 0.644) without;
the report-absent readouts answer \texttt{yes} on 73\% of trials, so the balanced accuracies are
the fair comparison.
\begin{table}[H]
\centering\small
\setlength{\tabcolsep}{4pt}
\adjustbox{max width=\textwidth}{%
\begin{tabular}{>{\raggedright\arraybackslash}p{0.37\textwidth}rrr}
\toprule
 & Report & Report & Absent minus \\
Statistic & present & absent & present \\
\midrule
Flip rate, 800 manual-label-changed trials & 3.13\% [1.87, 4.48] & 12.50\% [9.60, 16.13] & $+9.4$ points [6.1, 13.3] \\
Flip rate, 3{,}274 same-label trials & 4.40\% & & \\
Agrees with the substitute's manual label, 800 trials & 0.378 [0.342, 0.410] & 0.519 [0.487, 0.552] & $+0.141$ [0.100, 0.187] \\
Agrees with the source report's label, 800 trials & 0.616 [0.580, 0.651] & & \\
Agrees with the substitute's manual label, 150 trials, one-question design & 0.053 [0.022, 0.088] & 0.147 [0.087, 0.208] & \\
Agrees with the report, 717 trials whose substitute's manual label contradicts it & 0.633 [0.596, 0.671] & 0.481 [0.444, 0.518] & \\
The same 717 trials, one-question design & 0.979 [0.956, 1.000] & & \\
\bottomrule
\end{tabular}}
\caption{Readout under the manually annotated report labels of MIMIC-CXR-JPG, with and without
the report, all 14 questions unless stated. Brackets: 95\% patient-clustered percentile bootstrap
over the 90 covered patients, 2{,}000 draws; unbracketed cells have none.}
\label{tab:rad}
\end{table}

\section{Other models: MedGemma-4B\new{, a second family and a third}}\label{app:family}
MedGemma-4B, run through the same four one-question cells on the same 3{,}199 cases, shows the same
interaction with a larger effect. Its lowercase first-token readout changes on 3.16\% of cases with the report
[2.46, 3.92]. Without the report it changes on 19.01\% [16.64, 21.40]. The paired difference is
15.85 points [13.45, 18.25]. The ratio is 6.0 [4.8, 7.8]. Balanced accuracy on the original image
is 0.812 with the report and 0.810 without\new{, in both cases scored against that image's own
study's label. The 27B value in the report-absent cell of the same design is 0.479
(Table~\ref{tab:ablate}): without a report the 27B readout answers \texttt{yes} on 2{,}911 of the
3{,}199 cases, against 2{,}422 for 4B, so its specificity collapses to 0.06 while the 4B
specificity stays at 0.77. The two numbers describe the same image and the same cell in different
models, not the same model on different images}. On the all-14 unit the 4B rates are 7.42\% with the
report and 22.52\% without, a paired 15.1 points [13.8, 16.3], positive on all 14 findings. The
ratio is 3.0 [2.8, 3.3]. On the same 44{,}786 trials the 4B model flips 2.7 points more than the
27B model (Table~\ref{tab:head}).

Qwen3.5-9B~\citep{qwen2026qwen35} (a non-Gemma image-text model, revision c2022362) was run through \new{the one-question
design} on the same 3{,}199 cases, \new{report present and absent, both images,} with the same
prompt, its own chat template, thinking disabled and images capped at 802{,}816 pixels. Its
\new{token-family} readout sums the probabilities of each answer family (the log-sum-exp of the logits, never a sum of raw
logits) and takes the argmax between the two family probabilities at the first generated position;
the yes family is token ids 9405, 9175, 9542 and 7179 and the no family 2083, 2665, 874 and 2233,
the case and leading-whitespace variants of the two words in the Qwen3.5 vocabulary. \new{Without
an answer instruction this model answered \texttt{yes} on 3{,}197 of 3{,}199 report-absent cases, a
saturated prior under which the report-presence effect could not be established, and its
spontaneous first token was a bare yes or no on only 13 to 34\% of cases. With the instruction
``Answer with exactly Yes or No.'' appended, as in Appendix~\ref{app:readout}, every generation was
a bare Yes or No, and the generated and token-family readouts agree on 0.984 to 0.997 of cases, so
the readout is validated for this model. Under that instruction the report-presence effect
replicates in direction and significance. With the report the answer moves on 1.3\% of
substitutions [0.9, 1.7]. Without it the answer moves on 14.5\% [12.5, 16.9]. The paired difference
is 13.3 points [11.2, 15.6] and the ratio 11.6. With the report the answer agrees with the report's
label on 0.916 of cases. Without it the answer agrees with the substituted image's label at chance,
balanced accuracy 0.48, although the asked finding's label changes on 2{,}321 of the 3{,}199
substitutions. Image reading in this family is therefore weak, and the result is a replication of
report anchoring, not of image competence: the model follows the report when it is shown one and
moves with the image when it is not, without reading the image accurately.} One H100.

\begin{newpar}
\textbf{A third lineage and a scale check.} Two further models were run through the same
one-question design on the same 3{,}199 cases and 293 patients, under the instruction ``Answer with
exactly Yes or No.'', with the token-family and generated readouts side by side and a
patient-clustered percentile bootstrap of 2{,}000 draws at seed 0: LLaVA-NeXT on
Mistral-7B~\citep{liu2024llavanext,liu2023llava} (revision 2424fdd4), a lineage that is neither
Gemma nor Qwen, and Qwen3.5-27B, three times the size
of the Qwen3.5-9B arm above. The report-presence effect holds in both (Table~\ref{tab:families}):
the LLaVA-NeXT token-family readout moves on 2.59\% of substitutions with the report and 15.35\%
without, a paired 12.75 points [11.10, 14.44] and a ratio of 5.92, while Qwen3.5-27B moves on
0.88\% against 20.48\%, a paired 19.60 points [17.00, 22.40] and a ratio of 23.39; the generated
readout gives 12.91 points [11.25, 14.65] and 20.23 points [17.67, 22.92] on the same cases. Scale
does not shrink the effect: 19.60 points at Qwen3.5-27B against 13.3 at Qwen3.5-9B, with
MedGemma-27B at 0.84\% against 17.69\% under the same instruction. Both readouts are validated
here: format compliance is 1.000 in all eight cells, none of the 25{,}592 generations was
unscoreable, and the generated answer agrees with the token family on 0.9978 to 0.9987 of cases for
LLaVA-NeXT and 0.9881 to 0.9984 for Qwen3.5-27B. What replicates is report anchoring and not image
competence: with the report removed and the substituted image shown, agreement with that image's
own label is 0.518 (balanced 0.470 [0.452, 0.488]) for LLaVA-NeXT and 0.471 (balanced 0.516
[0.492, 0.539]) for Qwen3.5-27B\new{, with 0.487 [0.462, 0.511] for Qwen3.5-9B, patient-clustered
over 293 patients. The Qwen intervals include 0.5, which is insufficient evidence of above-chance
reading and not evidence of equivalence to chance; the LLaVA-NeXT interval sits below 0.5, which
on a label-conditioned set is consistent with following the source report rather than the
image}, both
chance, while balanced accuracy against the case's own report label is 0.925 and 0.943 with the
report present and falls to 0.670 and 0.848 once the report is removed. The MedGemma arms
complete the same cell: with the report removed and the substituted image shown, agreement with
that image's own label is balanced accuracy 0.480 for MedGemma-27B under the token-family readout
with this instruction (0.515 under the paper's lowercase readout, Appendix~\ref{app:readout}) and
0.518 [0.492, 0.542] for MedGemma-4B under the paper's readout without the instruction. No model
tested in this paper shows above-chance reading of the substituted image on this one-question set. That set
is label-conditioned, with target \texttt{yes} on 2{,}707 of 3{,}199 cases, and the statement does
not carry to the primary all-14 unit: there the report-absent readout reaches balanced accuracy
0.665 [0.658, 0.672] on the original image against a question-only cell at 0.500
(Table~\ref{tab:cells14}), and 0.666 [0.659, 0.674] on the substituted image against that study's
own asked-finding label. The at-chance statement is about this question set, not about MedGemma's
image reading in general.
\end{newpar}

\begin{newpar}
\textbf{Questions selected independently of the source label.} The one-question design asks a
finding-specific question on the 2{,}707 positive source cases and the composite question on the
492 negative ones, so a rule that answers yes to every finding-specific question and no to the
composite scores perfectly on the source label without reading anything, and source-image accuracy
on that design cannot establish competence. To remove the shortcut, the same four cells were rerun
under the answer instruction with generated answers on a prespecified set of finding-specific
questions asked of every case: the six findings whose source-label yes-rate lies between 20\% and
80\%, a rule fixed before any result was seen (atelectasis 22.3\%, cardiomegaly 27.1\%, edema
20.8\%, lung opacity 32.0\%, pleural effusion 32.4\%, support devices 32.7\%), 19{,}194 trials per
model over the 293 patients, 9{,}531 of them label-changing (Table~\ref{tab:families13q}). The
report effect survives the change of design in every model, at 13.1 to 20.9 points, the same order
as MedGemma's 16.7 under the same readout. Image-only balanced accuracy on the original image is
above chance for the Qwen models, 0.635 [0.620, 0.651] and 0.664 [0.650, 0.678], which the
label-conditioned design could not show, and near chance for LLaVA-NeXT, 0.538 [0.524, 0.553]; on
the substitute image all three sit at or just above chance, 0.497 to 0.537. The Qwen models'
original-image scores come with sensitivity near 0.9 and specificity 0.3 to 0.4, a yes-bias, so
above chance is not the same as competent. No question-only cell was run in this sweep, so that
baseline is absent here rather than approximated. Per-finding report effects are in the record.
\end{newpar}

\begin{table}[H]
\centering\small
\setlength{\tabcolsep}{4pt}
\adjustbox{max width=\textwidth}{%
\begin{tabular}{lrrrrrr}
\toprule
 & \multicolumn{2}{c}{Generated-answer flip rate} & Report effect & \multicolumn{2}{c}{Image-only balanced accuracy} & Follows sub. label \\
\cmidrule(lr){2-3}\cmidrule(lr){5-6}
Model & Report present & Report absent & points & Original image & Substitute image & no report \\
\midrule
Qwen3.5-9B & 6.04\% [5.5, 6.6] & 19.09\% [17.1, 21.4] & $+13.1$ [11.2, 15.1] & 0.635 [0.620, 0.651] & 0.532 [0.519, 0.545] & 52.69\% \\
Qwen3.5-27B & 6.39\% [5.9, 6.9] & 23.90\% [22.0, 25.9] & $+17.5$ [15.8, 19.3] & 0.664 [0.650, 0.678] & 0.537 [0.524, 0.551] & 49.69\% \\
LLaVA-NeXT-Mistral-7B & 4.33\% [3.9, 4.8] & 25.24\% [24.1, 26.3] & $+20.9$ [19.9, 21.9] & 0.538 [0.524, 0.553] & 0.497 [0.489, 0.505] & 44.85\% \\
\bottomrule
\end{tabular}}
\caption{\new{Cross-model results on six finding-specific questions selected by a prespecified
prevalence rule and asked of every case, 19{,}194 trials per model, generated answers under the
explicit instruction. Balanced accuracy is against each image's own report-derived label with no
report shown. The last column is the share of the 9{,}531 label-changing pairs whose no-report
answer on the substitute image follows the substitute's label. Brackets: 95\% patient-clustered
percentile bootstrap over 293 patients, 2{,}000 draws; the report effect resamples patients once
per draw and scores both conditions on that draw.}}
\label{tab:families13q}
\end{table}

\begin{table}[H]
\begin{newpar}[breakable=false]
\centering\small
\setlength{\tabcolsep}{4pt}
\adjustbox{max width=\textwidth}{%
\begin{tabular}{lrrr}
\toprule
Model & Report present & Report absent & Paired difference, points \\
\midrule
MedGemma-27B & 0.84\% [0.50, 1.27] & 17.69\% [15.88, 19.61] & $+16.9$ [15.0, 18.8] \\
Qwen3.5-9B & 1.3\% [0.9, 1.7] & 14.5\% [12.5, 16.9] & $+13.3$ [11.2, 15.6] \\
Qwen3.5-27B & 0.88\% [0.56, 1.22] & 20.48\% [17.95, 23.26] & $+19.60$ [17.00, 22.40] \\
LLaVA-NeXT Mistral-7B & 2.59\% [1.97, 3.31] & 15.35\% [13.65, 17.13] & $+12.75$ [11.10, 14.44] \\
\bottomrule
\end{tabular}}
\caption{Four models on one axis: one-question design, 3{,}199 cases and 293 patients, token-family
readout under the instruction to answer Yes or No, paired difference absent minus present in
points. Brackets: 95\% patient-clustered percentile bootstrap, 2{,}000 draws.}
\label{tab:families}
\end{newpar}
\end{table}

\section{Sensitivity of the primary effect to pairing and label choices}\label{app:cohort}
\begin{table}[H]
\centering\small
\setlength{\tabcolsep}{4pt}
\adjustbox{max width=\textwidth}{%
\begin{tabular}{lrrrr}
\toprule
 & & Report & Report & Difference, \\
Subset or weighting & Trials & present & absent & points \\
\midrule
All trials (primary) & 44{,}786 & 4.70\% [4.35, 5.06] & 17.07\% [15.78, 18.39] & $+12.4$ [11.2, 13.6] \\
Composite question excluded & 41{,}587 & 4.78\% [4.43, 5.16] & 18.14\% [16.83, 19.46] & $+13.4$ [12.2, 14.6] \\
Cross-patient pairs excluded & 44{,}520 & 4.70\% [4.34, 5.07] & 16.99\% [15.70, 18.26] & $+12.3$ [11.1, 13.4] \\
Same known view & 26{,}726 & 4.61\% [4.15, 5.10] & 15.21\% [13.85, 16.60] & $+10.6$ [9.3, 11.9] \\
Both views known & 38{,}444 & 4.80\% [4.43, 5.21] & 16.80\% [15.44, 18.20] & $+12.0$ [10.8, 13.3] \\
Known view mismatch & 11{,}718 & 5.23\% [4.68, 5.76] & 20.43\% [18.43, 22.49] & $+15.2$ [13.3, 17.2] \\
Equal-study weighting & 44{,}786 & 4.70\% [4.36, 5.06] & 17.01\% [15.71, 18.30] & $+12.3$ [11.2, 13.5] \\
Equal-patient weighting & 44{,}786 & 4.46\% [4.05, 4.89] & 18.75\% [17.62, 20.01] & $+14.3$ [13.2, 15.5] \\
\bottomrule
\end{tabular}}
\caption{Primary effect under alternative pairings, weightings and question sets, all 14 questions
unless stated, each subset resampled over its own patients. Same known view: AP to AP or PA to PA;
both views known: 453 blank-view pairs excluded. Brackets: 95\% patient-clustered percentile
bootstrap, \new{10{,}000 draws for the primary row and 2{,}000 for the
subsets}.}
\label{tab:sens}
\end{table}
\textbf{Label policies} (asked finding, both images, raw CheXpert values). With an uncertain label
read as negative (the paper), positive or excluded, the label-changed group is 9{,}347, 10{,}626 or
8{,}660 trials (44{,}786, 44{,}786 or 42{,}032 retained). The flip rate on label-changed trials with
the report present is 4.30\%, 4.65\% or 4.53\%, against 4.81\%, 4.72\% or 4.79\% on label-unchanged
trials. The changed-minus-unchanged difference is $-0.50$ points [$-1.03$, $0.02$] under the
paper's policy. It is $-0.07$ [$-0.56$, $0.42$] with uncertain read as positive. It is $-0.26$
[$-0.82$, $0.27$] with uncertain excluded. Concordant balanced accuracy is 0.717, 0.724 or 0.725,
and the mean aligned margin change is 0.35, 0.34 or 0.37. Treating unmentioned findings as excluded
rather than negative keeps 6{,}997 trials and gives $+0.63$ [$-0.17$, $1.48$]. Without the report
the label-changed trials flip less often than the unchanged ones under every policy (14.85\%
against 17.65\% for the paper's policy, $-2.80$ [$-4.22$, $-1.31$]).

\textbf{Composite question excluded.} The 13 finding-specific questions alone give 41{,}587 trials,
8{,}416 with the asked finding's label changed; the composite question contributes 931 of the 9{,}347
label changes. With the report, 27B concordant accuracy is 0.649, balanced accuracy 0.767 and
discordant accuracy 0.593 [0.579, 0.608]; the 4B values are 0.692, 0.791 and 0.624. The flip rate
on label-changed against label-unchanged trials is 4.4\% against 4.9\% for 27B and 7.2\% against
7.5\% for 4B. The 27B flip rate on these questions is 4.78\% with the report and 18.14\% without, a
paired difference of 13.36 points [12.16, 14.57] (Table~\ref{tab:sens}); for 4B it is 7.46\%
against 21.96\%, a difference of 14.51 points. The pooled balanced accuracies of the four modality
cells become 0.767, 0.655, 0.837 and 0.500 and the macro-averaged 0.718, 0.580, 0.776 and 0.500,
with interactions of $-0.226$ pooled and $-0.139$ macro-averaged [$-0.152$, $-0.124$]. The primary
headline keeps all 14 questions.

\begin{newpar}
\textbf{Substitute resampling.} The substitute image was redrawn under two further seeds on the
same 3{,}199 cases, reports, questions and concordant images, one-question design, report present
and absent, both images. With the report the flip rate was 1.38\%, 1.34\% and 1.31\% under seeds 0,
1 and 2, a spread of 0.06 points; without it 10.28\%, 10.00\% and 10.32\%, a spread of 0.31 points.
The paired effect of removing the report was 8.91, 8.66 and 9.00 points. A 95\% interval on that
effect which resamples both the patients and the substitute draw is [7.32, 10.65]. Which cases flip
changes with the substitute: with the report 66 cases flip under at least one seed and 19 under all
three, without it 554 and 139. The size of the report effect does not.
\end{newpar}

\begin{table}[H]
\centering\small
\setlength{\tabcolsep}{4pt}
\adjustbox{max width=\textwidth}{%
\begin{tabular}{lllrrrrr}
\toprule
Model & Label & Report & Correct, correct & Correct, wrong & Wrong, correct & Wrong, wrong & Total \\
\midrule
MedGemma-27B & Changed & present & 297 & 5{,}730 & 3{,}215 & 105 & 9{,}347 \\
MedGemma-27B & Changed & absent & 1{,}058 & 4{,}014 & 3{,}945 & 330 & 9{,}347 \\
MedGemma-27B & Unchanged & present & 22{,}073 & 929 & 774 & 11{,}663 & 35{,}439 \\
MedGemma-27B & Unchanged & absent & 14{,}428 & 3{,}085 & 3{,}171 & 14{,}755 & 35{,}439 \\
MedGemma-4B & Changed & present & 520 & 5{,}577 & 3{,}090 & 160 & 9{,}347 \\
MedGemma-4B & Changed & absent & 1{,}824 & 3{,}566 & 3{,}503 & 454 & 9{,}347 \\
MedGemma-4B & Unchanged & present & 22{,}560 & 1{,}484 & 1{,}159 & 10{,}236 & 35{,}439 \\
MedGemma-4B & Unchanged & absent & 16{,}641 & 3{,}832 & 3{,}975 & 10{,}991 & 35{,}439 \\
\bottomrule
\end{tabular}}
\caption{Correctness on the original then the substitute image, each against its own study's label
for the asked finding, all 14 questions. Concordant correct is columns one and two, discordant
correct one and three, flips one and four on changed rows and two and three on unchanged rows.}
\label{tab:transition}
\end{table}
Of the 9{,}347 label-changed trials, 61.3\% [60.2, 62.4] are correct on the original image and wrong
on the substitute. Conditional on being correct on the original, that share is 0.951 with the
report [0.943, 0.958]. Without the report it is 0.791 [0.766, 0.816]. On the 35{,}439
label-unchanged trials a change of correctness is exactly a flip: 1{,}703 trials change correctness
with the report and 6{,}256 without. Both accuracies of Table~\ref{tab:head} follow from the counts:
with the report, the 27B readout is correct on the original image on 29{,}029 of 44{,}786 trials
(0.648) and on the substitute on 26{,}359 (0.589), and the four flip cells sum to 2{,}105. Balanced
accuracy on the substitute is 0.633 [0.626, 0.640] for 27B and 0.625 for 4B. \new{Every accuracy
in this paper is scored on the asked finding's own label for the study whose image is shown, except
the accuracy columns of the text-control table, which are agreement with the source study's label
and are named as such there.}

\textbf{Margins.} Table~\ref{tab:margin} gives the margin statistics on the same 44{,}786 trials.
Per finding, the aligned change with and without the report: pleural effusion 0.52 and 3.94,
support devices 1.12 and 1.59, edema 0.32 and 2.45, pneumothorax 0.28 and 2.38, consolidation 0.18
and 2.31, cardiomegaly 0.28 and 0.99, the composite question 0.09 and 1.39; fracture, pleural other and
enlarged cardiomediastinum are near zero in both.
\begin{table}[H]
\centering\small
\setlength{\tabcolsep}{4pt}
\adjustbox{max width=\textwidth}{%
\begin{tabular}{>{\raggedright\arraybackslash}p{0.34\textwidth}rrr}
\toprule
 & Report & Report & Within-trial \\
Statistic & present & absent & difference \\
\midrule
Starting margin $m_c$ on the original image, quartiles & $-3.47$ / $-0.31$ / $8.23$ & $-1.13$ / $3.03$ / $7.92$ & \\
Flip rate, lowest / highest quartile of $|m_c|$ & 15.1\% / 0.15\% & 37.9\% / 4.3\% & \\
Mean absolute paired change & 0.690 [0.657, 0.726] & 2.307 [2.203, 2.415] & 1.62 [1.53, 1.70] \\
Aligned change, label-changed trials & 0.35 [0.29, 0.41] & 1.44 [1.26, 1.62] & \\
Aligned change, no-to-yes (4{,}637 trials) & 0.63 [0.53, 0.74] & & \\
Aligned change, yes-to-no (4{,}710 trials) & 0.07 [0.05, 0.09] & & \\
Aligned change, averaged over the 14 findings & 0.246 [0.199, 0.297] & 1.227 [1.050, 1.407] & \\
Absolute change above 0.5, share of trials & 32.2\% [30.8, 33.5] & 78.0\% [77.0, 79.0] & \\
\bottomrule
\end{tabular}}
\caption{Margin changes on the 44{,}786 all-14 trials. The absolute change is larger without the
report on 82.9\% of trials. Section~\ref{sec:res1} pools the aligned change over trials; the finding-averaged
row weights findings equally. Brackets: 95\% patient-clustered percentile bootstrap, 2{,}000 draws;
unbracketed cells have none.}
\label{tab:margin}
\end{table}

\section{Text-control hierarchy}\label{app:textctl}
One-question design, full arm, both images, MedGemma-27B. Differences are paired against the plain
full report on each condition's cases. The full-report row is a re-run of the one-question full arm
on the same 3{,}199 cases: it gives 1.38\% (44 flips) against the 1.41\% (45 flips) of
Table~\ref{tab:designs}\new{, a one-case difference between two runs of the same arm}. Removing only the sentences that name
the asked finding is not separated from removing the whole report ($+1.3$ points [$-0.3$, $2.7$]).
On the 1{,}702 cases carrying both sections, the findings section alone is within 0.3 points of the
full report [$-0.3$, $0.9$]. The impression alone is 4.4 points above it [3.0, 5.7]. Text-block
order on the same design: with the text first the flip rate is 2.84\% against 1.41\% image first
(Table~\ref{tab:designs})\new{; the per-arm counts are in Table~\ref{tab:sup-mimicarms}}.
\begin{table}[H]
\centering\small
\setlength{\tabcolsep}{4pt}
\adjustbox{max width=\textwidth}{%
\begin{tabular}{>{\raggedright\arraybackslash}p{0.27\textwidth}rrrr}
\toprule
 & & vs.\ full report, & Source-label & Source-label \\
Text shown with the image & Flip rate & points & agreement, original & agreement, substitute \\
\midrule
Full report & 1.38\% [0.96, 1.87] & reference & 0.931 [0.919, 0.941] & 0.930 [0.919, 0.940] \\
Findings section only (2{,}507 cases) & 1.44\% [1.01, 1.89] & $+0.2$ [$-0.2$, $0.6$] & 0.927 [0.915, 0.939] & 0.925 [0.913, 0.937] \\
Target sentences replaced by the opposite label & 2.53\% [1.87, 3.18] & $+1.2$ [0.3, 1.9] & 0.043 [0.034, 0.053] & 0.046 [0.037, 0.056] \\
One sentence stating the label & 3.13\% [2.42, 3.90] & $+1.8$ [0.9, 2.6] & 0.845 [0.819, 0.868] & 0.848 [0.822, 0.871] \\
Impression section only (2{,}394 cases) & 4.26\% [3.34, 5.31] & $+3.1$ [2.2, 4.1] & 0.873 [0.853, 0.891] & 0.866 [0.847, 0.883] \\
Clinical boilerplate, length matched & 7.06\% [5.90, 8.38] & $+5.7$ [4.5, 7.0] & 0.787 [0.759, 0.813] & 0.764 [0.736, 0.790] \\
Neutral prose, length matched & 8.22\% [6.81, 9.74] & $+6.9$ [5.4, 8.4] & 0.799 [0.771, 0.823] & 0.767 [0.737, 0.794] \\
Report removed & 10.28\% [8.72, 12.02] & $+8.9$ [7.5, 10.6] & 0.773 [0.744, 0.799] & 0.746 \\
Target sentences removed & 11.60\% [10.35, 12.93] & $+10.2$ [9.0, 11.6] & 0.743 [0.717, 0.767] & 0.726 [0.699, 0.751] \\
\bottomrule
\end{tabular}}
\caption{Text-control hierarchy, one-question design, 3{,}199 cases unless stated; accuracy is scored
against the source study's label. Sentences naming the asked finding were found by keyword lists (55
cases had none); section variants keep cases whose report has that section. Brackets: 95\%
patient-clustered percentile bootstrap, 2{,}000 draws; unbracketed cells have none.}
\label{tab:textctl}
\end{table}

\section{Readout validation}\label{app:readout}
This validation uses an explicit answer instruction and covers \new{both designs.} The one-question
design was re-run with and without the instruction ``Answer with exactly Yes or No.'' appended to
the question, eight tokens generated per case, report-present and report-absent arms, both images;
\new{the all-14 design was then rerun with the instruction (Table~\ref{tab:readout14}).} Three
readouts are compared: the paper's
lowercase first-token readout (argmax over the lowercase \texttt{yes}/\texttt{no} logits), the
token-family readout (argmax over the summed probabilities of the yes and no variants) and the
parsed generated answer. With the instruction, compliance is
100\%. Without it the model answers ``Yes'' only when the label is yes and otherwise opens with
``Based on the provided report'', so ``no'' answers are unscoreable within eight tokens.
\begin{table}[H]
\centering\small
\setlength{\tabcolsep}{4pt}
\adjustbox{max width=\textwidth}{%
\begin{tabular}{llrrr}
\toprule
Prompt & Arm & Lowercase first token & Token families & Generated answer \\
\midrule
With instruction & report present & 1.22\% [0.81, 1.67] & 0.84\% [0.50, 1.27] & 0.84\% [0.50, 1.27] \\
With instruction & report absent & 18.26\% [16.20, 20.57] & 17.69\% [15.88, 19.61] & 17.66\% [15.90, 19.55] \\
With instruction & paired difference & $-17.0$ [$-19.4$, $-14.9$] & $-16.9$ [$-18.8$, $-15.0$] & $-16.8$ [$-18.7$, $-15.0$] \\
No instruction & report present & 1.38\% [0.96, 1.87] & 1.09\% [0.69, 1.54] & 1.84\% [1.36, 2.37] \\
No instruction & report absent & 10.28\% [8.72, 12.02] & 17.35\% [15.31, 19.56] & 11.57\% [9.99, 13.32] \\
No instruction & paired difference & $-8.9$ [$-10.6$, $-7.5$] & $-16.3$ [$-18.5$, $-14.2$] & $-9.7$ [$-11.5$, $-8.1$] \\
\bottomrule
\end{tabular}}
\caption{Flip rates of the one-question design under three readouts and two prompts, 3{,}199 cases;
paired difference is report present minus absent, in points. No-instruction generated answers are
three-state (yes, no, unscoreable). Brackets: 95\% patient-clustered percentile bootstrap, 2{,}000
draws.}
\label{tab:readout}
\end{table}
Every answer is the argmax over two output logits at the first generated position, the single
tokens yes (id 4443) and no (id 1904) of the MedGemma-27B tokenizer at revision 2d3e00e, selected
as the unique one-token continuations of the chat-template generation prompt. The capitalised and
space-prefixed variants (Yes 10784, yes with a leading space 11262, Yes with a leading space 8438;
No 3771, no with a leading space 951, No with a leading space 2301) enter only the token-family
readout, where the model's top-1 first token is the capitalised form. MedGemma-4B at revision
290cda5e maps the same eight strings to the same ids. The exact tokens the paper reads carry a mean
first-position probability below $10^{-4}$ in every condition. Under the explicit instruction, the lowercase readout agrees with generated answers on
98.8\% of report-present cases and 91 to 93\% of report-absent cases. The instruction changes the
report-absent rate under the paper's readout by $+8.0$ points
but not under the family readout ($+0.3$ [$-1.4$, $2.0$]), so the report-absent value is a property
of the readout choice, and the report-presence effect is larger under the validated readouts. The
order of the two options is tested in Appendix~\ref{app:ansorder}.

\begin{newpar}
\textbf{The all-14 design under the instruction.} The primary design, 44{,}786 trials on the same
3{,}199 cases and 293 patients, was rerun with the instruction appended and greedy generation, so
the same trials can be scored by the generated answer and the token families as well as by the
lowercase first-token readout (Table~\ref{tab:readout14}). Format compliance, the share of
generations that open with Yes or No, is at least 0.9999 in every condition. The primary effect
survives on its own unit under every readout: the paired difference is 16.1 points under the
lowercase readout, 16.7 under the token families and 16.7 under the generated answer, with ratios
of 4.5 to 4.9, and every interval on the difference excludes zero. Under the headline prompt, which
carries no instruction, the same trials give 4.70\% against 17.07\%, a difference of 12.37 points;
the instruction raises the report-absent rate by about four points and leaves the report-present
rate within a tenth of a point under the lowercase readout, so the instructed prompt is a check on
the readout and not a replacement for the headline.
\end{newpar}
\new{Table~\ref{tab:readout14}, the all-14 design under the instruction, is in the main text (Section~\ref{sec:res1}).}

\section{Modality ablation on both designs}\label{app:cells14}
\begin{table}[H]
\centering\small
\setlength{\tabcolsep}{4pt}
\adjustbox{max width=\textwidth}{%
\begin{tabular}{>{\raggedright\arraybackslash}p{0.17\textwidth}rrrr}
\toprule
 & & Image + & Report + & Question \\
Finding & Full & question & question & only \\
\midrule
Atelectasis & 0.623 [0.603, 0.646] & 0.500 [0.500, 0.500] & 0.823 [0.807, 0.839] & 0.500 [0.500, 0.500] \\
Cardiomegaly & 0.766 [0.743, 0.788] & 0.656 [0.630, 0.682] & 0.808 [0.789, 0.827] & 0.500 [0.500, 0.500] \\
Consolidation & 0.834 [0.809, 0.858] & 0.645 [0.609, 0.685] & 0.887 [0.861, 0.909] & 0.500 [0.500, 0.500] \\
Edema & 0.849 [0.829, 0.868] & 0.658 [0.637, 0.681] & 0.882 [0.867, 0.897] & 0.500 [0.500, 0.500] \\
Enlarged cardiomediastinum & 0.480 [0.429, 0.530] & 0.482 [0.432, 0.533] & 0.511 [0.469, 0.559] & 0.500 [0.500, 0.500] \\
Fracture & 0.938 [0.903, 0.965] & 0.586 [0.532, 0.642] & 0.906 [0.857, 0.947] & 0.500 [0.500, 0.500] \\
Lung lesion & 0.696 [0.666, 0.725] & 0.499 [0.483, 0.512] & 0.772 [0.737, 0.804] & 0.500 [0.500, 0.500] \\
Lung opacity & 0.601 [0.583, 0.621] & 0.503 [0.501, 0.505] & 0.688 [0.668, 0.707] & 0.500 [0.500, 0.500] \\
Pleural effusion & 0.833 [0.811, 0.854] & 0.740 [0.712, 0.768] & 0.911 [0.897, 0.924] & 0.500 [0.500, 0.500] \\
Pleural other & 0.608 [0.492, 0.717] & 0.529 [0.428, 0.637] & 0.580 [0.475, 0.694] & 0.500 [0.500, 0.500] \\
Pneumonia & 0.714 [0.684, 0.742] & 0.537 [0.505, 0.569] & 0.710 [0.677, 0.742] & 0.500 [0.500, 0.500] \\
Pneumothorax & 0.853 [0.781, 0.903] & 0.711 [0.663, 0.758] & 0.830 [0.749, 0.889] & 0.500 [0.500, 0.500] \\
Support devices & 0.535 [0.525, 0.547] & 0.500 [0.500, 0.500] & 0.784 [0.757, 0.808] & 0.500 [0.500, 0.500] \\
Any finding (composite) & 0.732 [0.712, 0.750] & 0.524 [0.510, 0.543] & 0.696 [0.680, 0.712] & 0.500 [0.500, 0.500] \\
\midrule
Pooled balanced accuracy & 0.717 [0.710, 0.725] & 0.665 [0.658, 0.672] & 0.770 [0.764, 0.776] & 0.500 [0.500, 0.500] \\
Macro-averaged balanced accuracy & 0.719 [0.706, 0.732] & 0.576 [0.564, 0.588] & 0.770 [0.759, 0.782] & 0.500 [0.500, 0.500] \\
Pooled micro accuracy & 0.648 [0.635, 0.662] & 0.504 [0.493, 0.516] & 0.772 [0.764, 0.781] & 0.204 [0.194, 0.213] \\
Sensitivity & 0.834 [0.817, 0.847] & 0.937 [0.926, 0.945] & 0.767 [0.751, 0.781] & 1.000 [1.000, 1.000] \\
Specificity & 0.601 [0.580, 0.623] & 0.394 [0.374, 0.415] & 0.773 [0.760, 0.787] & 0.000 [0.000, 0.000] \\
\bottomrule
\end{tabular}}
\caption{Balanced accuracy per finding, pooled and macro-averaged, four modality cells, all-14
design (44{,}786 trials, report-derived labels, majority 0.796). Interaction, image effect with the
report minus without: $-0.218$ [$-0.224$, $-0.212$] pooled over trials. Macro-averaged over
findings: $-0.128$ [$-0.141$, $-0.115$]. Brackets: 95\% patient-clustered percentile bootstrap,
2{,}000 draws.}
\label{tab:cells14}
\end{table}

\begin{table}[H]
\centering\small
\setlength{\tabcolsep}{4pt}
\adjustbox{max width=\textwidth}{%
\begin{tabular}{>{\raggedright\arraybackslash}p{0.21\textwidth}rrrrrr}
\toprule
 & \multicolumn{2}{c}{Full} & \multicolumn{2}{c}{Image + question} & \multicolumn{2}{c}{Report + question} \\
\cmidrule(lr){2-3}\cmidrule(lr){4-5}\cmidrule(lr){6-7}
Finding & Sens. & Spec. & Sens. & Spec. & Sens. & Spec. \\
\midrule
Atelectasis & 0.999 & 0.247 & 1.000 & 0.000 & 0.993 & 0.652 \\
Cardiomegaly & 0.893 & 0.639 & 0.879 & 0.432 & 0.833 & 0.783 \\
Consolidation & 0.957 & 0.711 & 0.791 & 0.499 & 0.909 & 0.865 \\
Edema & 0.968 & 0.730 & 0.956 & 0.361 & 0.958 & 0.806 \\
Enlarged cardiomediastinum & 0.406 & 0.554 & 0.594 & 0.370 & 0.219 & 0.804 \\
Fracture & 0.925 & 0.950 & 0.280 & 0.893 & 0.839 & 0.972 \\
Lung lesion & 0.952 & 0.441 & 0.976 & 0.022 & 0.911 & 0.633 \\
Lung opacity & 0.991 & 0.212 & 1.000 & 0.005 & 0.969 & 0.407 \\
Pleural effusion & 0.990 & 0.675 & 0.908 & 0.572 & 0.988 & 0.833 \\
Pleural other & 0.694 & 0.521 & 0.444 & 0.613 & 0.486 & 0.673 \\
Pneumonia & 0.679 & 0.749 & 0.746 & 0.327 & 0.560 & 0.861 \\
Pneumothorax & 0.723 & 0.983 & 0.607 & 0.814 & 0.670 & 0.991 \\
Support devices & 0.994 & 0.077 & 1.000 & 0.000 & 0.973 & 0.594 \\
Any finding (composite) & 0.589 & 0.876 & 0.994 & 0.055 & 0.444 & 0.947 \\
\midrule
Pooled & 0.834 & 0.601 & 0.937 & 0.394 & 0.767 & 0.773 \\
\bottomrule
\end{tabular}}
\caption{\new{Sensitivity and specificity per finding, three modality cells, all-14 design
(44{,}786 trials, report-derived labels). Question only is omitted: it answers yes on every
trial, so its sensitivity is 1.000 and its specificity 0.000 for every finding. These are the
decomposition of the balanced accuracies in Table~\ref{tab:cells14}, which is their mean, and
they are point estimates from the committed per-finding summary with no intervals computed.
The decomposition is what the balanced accuracy hides: on the full arm, atelectasis, lung
opacity and support devices sit at 0.999, 0.991 and 0.994 sensitivity against 0.247, 0.212 and
0.077 specificity, so a balanced accuracy near 0.6 for those findings is a near-constant yes
rather than a discrimination. Removing the image raises specificity on all 14 findings,
which is the same ordering the pooled row shows.}}
\label{tab:sensspec14}
\end{table}

\begin{table}[H]
\centering\small
\setlength{\tabcolsep}{5pt}
\begin{tabular}{lrrr}
\toprule
Finding & Report present & Report absent & Absent minus present \\
\midrule
Atelectasis & 6.81\% & 4.63\% & $-2.2$ \\
Cardiomegaly & 6.06\% & 26.23\% & $+20.2$ \\
Consolidation & 3.84\% & 30.73\% & $+26.9$ \\
Edema & 2.50\% & 28.63\% & $+26.1$ \\
Enlarged cardiomediastinum & 4.78\% & 24.29\% & $+19.5$ \\
Fracture & 0.78\% & 7.00\% & $+6.2$ \\
Lung lesion & 6.19\% & 26.95\% & $+20.8$ \\
Lung opacity & 5.10\% & 12.38\% & $+7.3$ \\
Pleural effusion & 5.22\% & 29.60\% & $+24.4$ \\
Pleural other & 2.31\% & 21.76\% & $+19.5$ \\
Pneumonia & 2.50\% & 33.17\% & $+30.7$ \\
Pneumothorax & 0.31\% & 12.32\% & $+12.0$ \\
Support devices & 11.03\% & 1.44\% & $-9.6$ \\
Any finding (composite) & 2.13\% & 34.10\% & $+32.0$ \\
\midrule
Pooled, 14 questions & 4.26\% & 20.94\% & $+16.7$ [15.6, 17.7] \\
Pooled, 13 finding-specific & 4.42\% & 19.93\% & $+15.5$ [14.5, 16.5] \\
\bottomrule
\end{tabular}
\caption{\new{Image-swap flip rate of the generated answer per finding under the explicit answer
instruction, all-14 design, 3{,}199 cases per finding. Two findings reverse the pooled direction:
atelectasis and support devices flip less often without the report than with it. The pooled
increase is therefore a positive pooled effect, not a consistent effect across findings. Intervals
on the pooled rows are patient-clustered, 10{,}000 draws; per-finding rows are point estimates
from the committed per-trial record.}}
\label{tab:genperfinding}
\end{table}
Table~\ref{tab:ablate} gives the same four cells on the secondary one-question design.
\begin{table}[!htb]
\centering\small
\setlength{\tabcolsep}{4pt}
\adjustbox{max width=\textwidth}{%
\begin{tabular}{lccccc}
\toprule
Arm & Report & Image & Accuracy & Balanced acc. & $\Delta$ vs.\ full, paired \\
\midrule
Full (reference) & \checkmark & \checkmark & 0.931 [0.919, 0.941] & 0.908 & reference \\
Report + question & \checkmark & $\times$ & 0.913 [0.900, 0.925] & 0.927 & $-0.018$ [$-0.026$, $-0.009$] \\
Question only & $\times$ & $\times$ & 0.846 [0.822, 0.869] & 0.500 & $-0.084$ [$-0.109$, $-0.062$] \\
Image + question & $\times$ & \checkmark & 0.773 [0.744, 0.799] & 0.479 & $-0.158$ [$-0.184$, $-0.134$] \\
\bottomrule
\end{tabular}}
\caption{Modality ablation, one question per case (3{,}199 cases, 293 patients), question always
present. Target \texttt{yes} on 2{,}707 cases and \texttt{no} on the 492 asked the composite question, so the
majority rate 0.846 equals question-only accuracy. \new{Both accuracy columns are on the original
image, scored against its own study's label; the MedGemma-4B counterparts are in
Appendix~\ref{app:family}.} Brackets: 95\% patient-clustered percentile
bootstrap, 10{,}000 draws; unbracketed cells have none.}
\label{tab:ablate}
\end{table}

\section{Classifier-surrogate image labels}\label{app:clf}
As a surrogate for image-level labels, not a reading, a DenseNet-121 chest X-ray classifier from
TorchXRayVision~\citep{cohen2022torchxrayvision} version 1.5.4, the checkpoint the library names
densenet121-res224-chex, trained on CheXpert only~\citep{irvin2019chexpert} and so on neither these
images nor these reports, scored all 3{,}300 shown images on CPU with the library's own
preprocessing: each image is read as 8-bit grayscale, mapped linearly to the library's input range
of $-1024$ to $1024$, centre-cropped to a square on its long side and resized to 224 by 224 pixels
with anti-aliasing, with no other normalisation. Eleven of its output heads map one to one onto
atelectasis, cardiomegaly, consolidation, edema, enlarged cardiomediastinum, fracture, lung lesion,
lung opacity, pleural effusion (the head named Effusion), pneumonia and pneumothorax; the composite
any-finding question takes the maximum of those eleven scores; pleural other and support devices
have no head and are unlabelled on every trial, so the analysis covers twelve questions per case,
38{,}388 trials. Thresholds were applied not to the raw sigmoid probability but to the library's
operating-point-normalised score, a piecewise-linear rescaling under which each head's
CheXpert-fitted operating point (a raw probability between 0.053 for pneumothorax and 0.327 for
lung opacity) sits at 0.5: an image is called present at or above 0.5 and absent below, and a trial
is excluded whenever either of its two images scores within 0.1 of 0.5 on that scale, a band that
on the raw probability scale is always 0.2 wide and starts at 0.8 times the operating point (0.16
to 0.36 for atelectasis, for example). The rule was fixed before any flip was examined. Of the
38{,}388 trials, 2{,}249 carry a classifier-predicted label change of the asked finding between the
two images, 22{,}469 a classifier-predicted label preservation and 13{,}670 fall in the excluded
band; the 6{,}398 trials on pleural other and support devices lie outside this denominator
(Table~\ref{tab:clf}). Against the report-derived labels on the 3{,}199 source images the
surrogate's AUROC runs from 0.53 (pneumonia) to 0.74 (pleural effusion) across the eleven heads and
is 0.76 for the derived any-finding score; at the 0.5 rule it calls between 18\% (pneumothorax) and
92\% (atelectasis) of images positive, above one half for ten of the eleven heads, against
report-label prevalences of 3\% to 32\%. Re-applying the same rule to the raw sigmoid probability
instead (present at or above 0.5, excluded between 0.4 and 0.6) leaves every AUROC unchanged and
keeps 29{,}147 trials (4{,}706 with a predicted change, 24{,}441 with a predicted preservation,
9{,}241 excluded). Under that rule the flip rate with the report is 7.44\% on predicted changes
[6.51, 8.37] against 3.64\% on predicted preservations. Without the report it is 31.24\%
[28.22, 34.09] against 13.82\%, the same direction as the primary rule.
\begin{table}[H]
\centering\small
\setlength{\tabcolsep}{4pt}
\adjustbox{max width=\textwidth}{%
\begin{tabular}{>{\raggedright\arraybackslash}p{0.30\textwidth}rrr}
\toprule
 & & Agrees with & Paired report \\
Report and classifier-predicted subset & Flip rate & substitute label & difference, points \\
\midrule
Present, label change (2{,}249) & 7.20\% [6.03, 8.57] & 0.422 [0.395, 0.451] & reference \\
Present, label preservation (22{,}469) & 4.44\% [3.99, 4.87] & 0.697 [0.678, 0.715] & reference \\
Absent, label change & 37.35\% [32.63, 41.87] & 0.659 [0.624, 0.689] & $+30.1$ [25.3, 34.6] \\
Absent, label preservation & 12.06\% [10.98, 13.17] & 0.846 [0.831, 0.861] & $+7.6$ [6.6, 8.7] \\
Absent, change supported by classifier and report labels (396) & 51.01\% [41.35, 58.41] & 0.745 [0.686, 0.790] & $+44.9$ [36.1, 52.2] \\
\bottomrule
\end{tabular}}
\caption{Flip rates where the classifier surrogate predicts a change in the queried finding or the
same label for both images; last row: change supported by both classifier and report labels.
Brackets: 95\% patient-clustered percentile bootstrap, 2{,}000 draws; 293 patients on the
classifier-only rows, 286 on the last.}
\label{tab:clf}
\end{table}
Where the surrogate predicts a change in the queried finding, the readout without the report flips
three times as often as where it predicts the same label for both images; the report-derived split
shows no such separation on the same findings (4.55\% against 4.69\% with the report, 16.91\%
against 17.68\% without).

\section{A second question template}\label{app:template}
The one-question design was re-run with the template ``Is <finding> present on this chest X-ray?''
(``Is any acute cardiopulmonary finding present on this chest X-ray?'' for the composite question) in place of
``Does this chest X-ray show <finding>?'', report present and absent, both images, no answer
instruction. The flip rate is 1.06\% [0.70, 1.42] with the report. It is 8.28\% [6.55, 10.15]
without, against 1.38\% and 10.28\% under the default template on the same cases. The
report-presence paired difference is $-7.2$ points [$-9.2$, $-5.5$] under the second template. It is
$-8.9$ [$-10.6$, $-7.5$] under the default. The interaction is $+1.7$ [$-0.0$, $+3.4$]. The second
template raises report-absent balanced accuracy from 0.479 to 0.621 (specificity 0.06 to 0.29)
without changing the report-present value (0.908).

\section{Answer order in the instruction}\label{app:ansorder}
The four passes carrying the answer instruction were repeated with the two options reversed, from
``Answer with exactly Yes or No.'' to ``Answer with exactly No or Yes.'', paired case by case with
the default order on the same 3{,}199 cases. Under the paper's readout the report-present flip rate
is 0.97\% reversed against 1.22\% default, a difference of $-0.25$ points [$-0.60$, $0.09$]. The
report-absent rate is 18.54\% against 18.26\%, a difference of $+0.28$ points [$-0.42$, $0.99$]. The
report-presence difference is $-17.57$ points under the reversed order and $-17.04$ under the
default; the gap between the orders is $-0.53$ points [$-1.36$, $0.30$]. Under the generated answer
the reversed order raises the report-absent flip rate by 1.22 points [0.51, 1.90], and by 1.09
points under the token families, an order of magnitude below the report effect. Reversing the order
changed the answer on 0.59\% to 1.78\% of cases and lowered the yes rate by 0.47 to 1.03 points with
the report and by up to 1.97 points under generation, a small shift toward the option listed first.

\textbf{An explicit abstention option.}\label{app:abstain}
Rerun on the same 3{,}199 cases with the model allowed to answer Yes, No or Uncertain, the
report-absent condition never produced Uncertain on either image, and the report-present condition
did so on 1.5\% of cases [1.0, 2.0]. The image substitution left that rate unchanged, $+0.06$ points
[$-0.10$, $0.22$] substituted minus original. Counting a change to or from Uncertain as a change,
the answer flipped with the substituted image on 1.1\% of cases with the report and 18.3\% without,
a paired difference of 17.2 points [15.2, 19.3]. Under the forced yes-or-no instruction the same
difference was 16.8 points [15.0, 18.7]. An explicit abstention option therefore neither weakens
nor explains the report effect: the model abstains only when a report is present, and at the same
rate whether or not the image agrees with it.

\section{\new{Layer-wise readouts and attention interventions}}\label{app:neg}
\begin{newpar}
This appendix asks how the measured response to an image substitution evolves across decoder
depth, and what an attention intervention adds to that description. We evaluated four approaches
to characterising the effect across decoder layers.

\emph{Reading the answer off each layer} fails, and it fails for a measurable reason. The
projection does not track the model's own answer until layer 46 of 62, so at every depth where the
interesting change would happen it is reporting the lens rather than the network. \emph{The tuned
lens}, the standard remedy, scored below the uncalibrated logit lens on both readout datasets, so
the fix does not rescue the instrument. \emph{The settling depth}, the first layer at which the
projected answer stops changing, is the most direct form of the question and gives one-question
medians of 26 with the report and 23 without, but a shuffled-pairing null reproduces the same
per-case difference, so the statistic carries nothing about the pairing. \emph{A per-layer probe}
transfers to chance under a frozen protocol on a disjoint cohort.

\emph{Attention knockout} provides an intervention-based comparison. Blocking the report's influence on the text stream restores
image sensitivity when the block starts shallow and does nothing when it starts deep, which
measures the effect of access to report-token representations by depth. It is an intervention
rather than a reading off a projection, which is what the first three approaches could not
deliver, and it does not localise report information that has already left those positions. Cross-condition patching adds no localisation on top of it.

The order below follows that argument: the failed readout first, with the calibration results that
limit its interpretation, then the intervention result, then the two negative interventions that motivated the
question. Nothing here rests on the layerwise projection.
\end{newpar}

\subsection{\new{The projection and its calibration}}
\begin{figure}[H]
\centering
\includegraphics[width=0.55\textwidth]{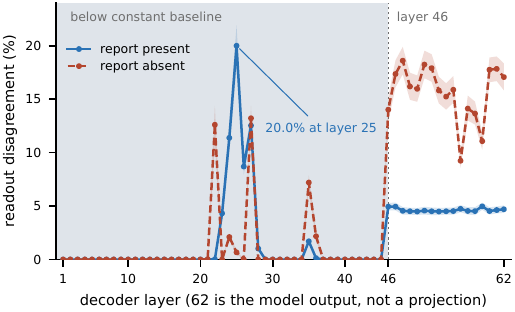}
\begin{newpar}[breakable=false]
\caption{Uncalibrated logit-lens projection: share of 44{,}786 trials, 14 findings pooled, whose
projected yes/no readout differs between the two images, by layer. Below layer 46 (shaded)
the projection does not clear the best constant predictor of the model's output. The
layer-25 peak, 20.0\% with the report against 0.7\% without, is a property of the report-present
prompt rather than of the network and supports no claim.}
\label{fig:layer}
\end{newpar}
\end{figure}

Both analyses are exploratory and negative; neither supports a mechanistic claim. The layerwise
logit-lens projection~\citep{nostalgebraist2020logitlens} (at each of the 62 decoder layers, the
fraction of report-present trials whose projected readout differs between the two runs) describes
the projection only, and no accuracy, flip rate or margin change depends on it. Its output-layer
value is the flip rate of Table~\ref{tab:head}, the per-split values are in Appendix
Table~\ref{tab:sup-settling}, and it is not evidence of where a decision is made
(Figure~\ref{fig:layer}). Appendix Table~\ref{tab:sup-mimicarms} also counts pairs still apart at
the final layer: flips count pairs whose final answer differs between the two images, while still
apart counts pairs whose projected yes/no distributions at the final layer differ by more than 0.05
in total variation, whichever answer wins, so it exceeds the flip count (48 against 6 with the
image first, 122 against 33 with the text first) and neither count implies the other.

\begin{newpar}
A calibration pass on the cached readouts asked, at each of the 62 positions, whether the
lens-projected yes/no answer agrees with the model's own forced-choice output on the same trial,
balanced over the two output classes and scored against the best constant predictor of that
output. The projection clears both baselines only from layer 46 of 62, and layer 46 is the
shallowest such depth in all 2{,}000 patient-clustered draws and in all eight arms (95\% CI 46 to
46); balanced agreement there is 0.909 [0.894, 0.924] with the report and 0.760 [0.729, 0.790]
without it. Below that depth the projection carries a limited and non-persistent signal rather than none: against a constant predictor, whose balanced agreement is 0.500, it reaches 0.648 at layer 25 and falls back; it emits a single constant answer at
37 of the 61 projected layers, and after a brief excursion at layer 25, where balanced agreement
reaches 0.648 [0.628, 0.668], it falls back to within 0.002 of 0.500 at every layer from 28 to 45.
The layer-25 feature is also prompt-specific: on the all-14 unit, readout disagreement at layer 25
is 0.2000 [0.179, 0.219] with the report against 0.0067 [0.005, 0.008] without it, a difference of
19.3 points [17.2, 21.3], and the report-absent curve rises only after layer 46. The settling
depth, the first layer from which the projected answer stops changing, has one-question medians of
26 with the report and 23 without, a paired mean difference of 1.56 layers [0.88, 2.21] against a
sign-flip null centred at zero, but a shuffled-pairing null reproduces the same median per-case
difference of 3.0, so the pairing carries nothing and the statistic is a difference of marginal
distributions read some 20 layers shallower than layer 46. \new{No projected claim in this paper is anchored to a depth below layer 46; the only depth
statement below it is the causal bound from the attention knockout above, which does not use the
lens.} The layer-25 peak is a property of the report-present prompt rather than of the network, and the final entry of each stored row is the model's own output logits rather than a
projection, which is why the curve's endpoint equals the flip rate exactly, 0.0470 [0.044, 0.051]
with the report and 0.1707 [0.158, 0.183] without.
\end{newpar}

\subsection{\new{Report-token attention knockout}}
\begin{figure}[t]
\centering
\includegraphics[width=0.55\textwidth]{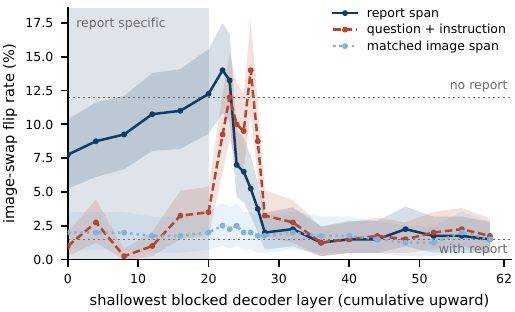}
\begin{newpar}[breakable=false]
\caption{Report-token attention knockout, cumulative from the plotted layer upward, wider query
scope, 400 one-question cases. Blocking the report restores image sensitivity toward the no-report
rate only when the block starts at or below layer 22; the question-and-instruction control moves
only over layers 22 to 27. Shading: layers 0 to 20.}
\label{fig:knockout}
\end{newpar}
\end{figure}

\begin{newpar}
\textbf{Causal depth: report-token attention knockout and cross-condition patching.} Two
interventions measure the effect of access to report-token representations, on MedGemma-27B and the
one-question design. The first masks attention at decoder layer $L$ so that chosen query positions
cannot read the report's key columns, cumulatively from $L$ upward or at $L$ alone, with the query
scope either the answer position alone or every non-image position outside the blocked span. It
covers 400 cases and 282 patients, 300 of them decisive, at 21 of 62 layers, with a 2{,}000-draw
patient-clustered bootstrap.

Under the cumulative variant and the wider scope the flip rate is 7.75\% [5.25, 10.35] from layer
0, 12.25\% from layer 20 and a peak of 14.00\% at layer 22, then 2.00\% at layer 28 and baseline
from layer 32, against in-run endpoints of 1.50\% with the report and 12.00\% without
(Figure~\ref{fig:knockout}). Single-layer blocking never exceeds 2.50\% and answer-position-only
blocking never exceeds 2.00\%, so no one layer carries the effect. Blocking the question and the
instruction instead also restores flips, but only over layers 22 to 27, so there a flip rate cannot
separate a restored image answer from one destroyed by not knowing what was asked.

Three readings confine the report-specific claim to layers 0 to 20. The paired patient-clustered
difference clears zero against both controls at layers 0, 4, 8, 12, 16 and 20, on all cases and on
the decisive subset, at layer 16 by $+7.75$ points [4.67, 11.03]. The arms flip disjoint cases over
layers 0 to 27 (Jaccard 0.000 to 0.069, $\phi$ $-0.076$ to $+0.079$) but the same cases in the
inert region at layers 32 to 60 (Jaccard 0.250 to 0.500, $\phi$ $+0.387$ to $+0.662$), so the
statistic can see a shared effect and sees none here. Their destinations differ: over layers 0 to
20 the report arm holds report-label agreement on the substituted image at 0.6875 to 0.7400, near
the measured no-report 0.7100, while the question-and-instruction arm drives it to 0.1725 to
0.2125, not an operating point at all. The span controls behave, a magnitude-matched image span
flat at 0.0125 to 0.0250 at every depth and a report span size-matched to the shorter non-report
text peaking at 4.75\%, a dose dependence on how much of the report is blocked.

Two limits bound the reading. The instrument has a ceiling, since the answer position still reads
the report directly and the report's tokens still occupy their positions and shift every later
rotary position, so blocking cumulatively from layer 0 yields 7.75\% against a no-report 12.00\%, and the intervention's extent should not be read as its effect size. Layer 22
over-restores, 14.00\% against that 12.00\%, with its agreement destination unchanged at 0.7075, so
the excess is added disruption rather than more removal. The second intervention patches the
report-absent residual into the report-present run, at the answer position and over the two
prompts' longest common token suffix, against a wrong-donor control from a different patient. It is
a null: the wrong donor reaches 14.07\% [10.78, 17.65], above the correct donor's maximum of
12.25\%, and no depth separates from its control. Cumulative blocking of attention to report-token positions increased image-swap sensitivity
when initiated at layers 0 to 20 and diminished for blocks initiated later. That identifies when
direct access to report-token positions affects the measured response; it does not localise all
downstream use of report information, since content copied out of those positions at an early
layer is untouched by a later block, small single-layer effects do not show that no layer carries
the effect, and a higher flip rate is not by itself a restored reading of the substituted image.
Patching adds no localisation.
\end{newpar}

\subsection{\new{Interventions that did not work: steering and probe transfer}}
\textbf{Steering.} Steering is scored on 1{,}721 test trials (12 findings, 150 cases each, the trials
whose answer was the same on both images) by whether the steered answer on the substituted image
matches that image's answer key; this is an accuracy against the substituted key, not the flip rate
of Table~\ref{tab:head}. Without steering, 835 of 1{,}721 (0.485) already match, because the
substituted key often equals the original one. Adding the difference-of-means
direction~\citep{marks2023geometry,turner2023actadd,li2023iti,rimsky2024caa} at decoder layer 20 at
the largest scale (6{,}000) changes this by two trials (837, 0.486), while injecting it at every
layer or injecting a random direction of the same norm moves it down by 18 and 14 trials;
degradation, the share of 1{,}209 originally correct answers that become wrong, is 0.167, 0.614 and
0.328. No arm separates from the random direction below the scale that degrades the answer. The
prompt is the main experiment's and the readout is the model output, not a layer projection; only
the largest scale is shown, the direction, layer choice and magnitude are as run, not normalised to
activation norms, and the record carries no confidence intervals (Table~\ref{tab:steer}).

\begin{table}[H]
\centering\small
\setlength{\tabcolsep}{4pt}
\adjustbox{max width=\textwidth}{%
\begin{tabular}{lccc}
\toprule
Condition & Tracks-substitute rate & Degradation rate & $\Delta$ vs.\ scale 0 \\
\midrule
Targeted, layer 20  & 0.486 (837/1{,}721) & 0.167 & $+0.001$ \\
Global, all layers  & 0.475 (817/1{,}721) & 0.614 & $-0.010$ \\
Random direction    & 0.477 (821/1{,}721) & 0.328 & $-0.008$ \\
\bottomrule
\end{tabular}}
\caption{Difference-of-means steering on MIMIC-CXR, 12 findings, 150 cases each, largest scale
(6{,}000). Tracks-substitute rate: share of 1{,}721 same-answer trials whose steered answer on the
substitute matches its answer key, 0.485 before steering; not the flip rate of
Table~\ref{tab:head}. Degradation: share of 1{,}209 originally correct answers turned wrong. No
intervals.}
\label{tab:steer}
\end{table}

\textbf{Probe transfer.} A linear probe~\citep{alain2016probes} was fit under a protocol frozen
before the cohort was sampled, then run once on 650 patients disjoint from every patient used
anywhere else in this study. The protocol is a configuration file whose SHA-256 hash (beginning
a756e332) is recorded in every output of the run; it fixes the probe layer (20), the four probe
findings, the split rule (650 patients drawn with seed 0 from the training pool, disjoint from the
293 test patients and from every patient in any cached store, 4{,}824 excluded, 1{,}811 available,
checked in code), the six baseline values with their sources, and the test (two-sided,
$\alpha{=}0.05$, patient-clustered bootstrap with 10{,}000 resamples, equality rejected when the
interval on the difference excludes zero), and it allows one run with no re-runs, added arms or
filters. \new{What the repository history verifies is narrower than a timestamp before the draw: the
protocol hash a756e332 appears in every output of the run, which started on 2026-08-04, while the
protocol file itself entered version control on 2026-09-03. The protocol was fixed before
evaluation; that it was fixed before the cohort was drawn is the authors' statement, not something
the history can establish.}
Test-split values are the frozen baselines, case-weighted over 13 findings, hence 0.952 against the
14-finding 0.953 of Appendix Table~\ref{tab:floor}. Every row of Table~\ref{tab:frozen} rejects
equality; the probe AUROC falls by 0.334, to a value not separable from chance.

\begin{table}[H]
\centering\small
\setlength{\tabcolsep}{4pt}
\adjustbox{max width=\textwidth}{%
\begin{tabular}{lccc}
\toprule
Statistic & Fresh cohort ($n{=}650$) & Test split & Gap (fresh $-$ test) \\
\midrule
Unchanged-answer rate   & 93.95\% [93.53, 94.37] & 95.22\% & $-1.3$ [$-1.7$, $-0.9$] \\
Never-diverged rate     & 57.10\% [55.63, 58.47] & 62.30\% & $-5.2$ [$-6.7$, $-3.8$] \\
Settling layer, median  & 61.0 [61.0, 61.0]    & 60.8  & $+0.2$ [0.2, 0.2] \\
Probe AUROC, layer 20   & 0.458 [0.414, 0.512] & 0.792 & $-0.334$ [$-0.379$, $-0.280$] \\
AURC~\citep{geifman2017selective}, probe & 0.414 [0.390, 0.437] & 0.490 & $-0.076$ [$-0.099$, $-0.053$] \\
AURC, max-softmax       & 0.346 [0.323, 0.369] & 0.374 & $-0.028$ [$-0.051$, $-0.005$] \\
\bottomrule
\end{tabular}}
\caption{Frozen-protocol replication on a disjoint MIMIC-CXR cohort; the pre-specified two-sided
test at $\alpha{=}0.05$ rejects equality in every row. Baselines are case-weighted over 13 findings
(rates) or the four probe findings (AUROC, AURC) and are fixed constants. Brackets: 95\%
patient-clustered percentile bootstrap, 10{,}000 draws; unbracketed cells have none. \new{The
settling-layer row is a property of the projection, whose median sits well above the layer-46
informative depth in this arm and some 20 layers below it on the one-question unit, where the
paired report-present minus report-absent difference is 1.56 layers [0.88, 2.21] and a
shuffled-pairing null reproduces it; the row supports no depth claim.}}
\label{tab:frozen}
\end{table}

\subsection{\new{The knockout does not transfer as a depth instrument outside MIMIC-CXR}}
\begin{newpar}
The report-token knockout above is possible only on MIMIC-CXR, the one dataset here that pairs an
image with a report. The nearest manipulation available elsewhere blocks attention to the
\emph{image} tokens instead, and it was run on three further datasets. It does not separate from
its own control on any of them, so no cross-dataset depth claim is made. On VQA-RAD ($n{=}52$)
blocking every image token changes the answer on 48.1\% of cases [34.6, 61.5], and blocking a
count-matched set of random tokens changes it on 51.9\% [38.5, 65.4]: the control is not below the
treatment. On OmniMedVQA the image block changes 8.0\% of answers [5.6, 10.8] against 83.6\%
[80.1, 86.9] for the random-key control, which is not a control that has been matched to the
manipulation at all. On SLAKE ($n{=}186$) the image block changes 36.6\% [29.6, 43.6] and no control
was run; its record marks the per-layer sweep exploratory and not a depth claim, and it is not read
as one here. Two of these three records also fail the repository's run-record schema and are
listed among its grandfathered exceptions. The negative is reported because it bounds the paper:
the depth result rests on the report-token knockout and its two controls on MIMIC-CXR, and on
nothing else.
\end{newpar}


\section{Cohort and labels}
Arm: the manipulation a row applies. Flip rate: share of pairs whose final answer differs between the original and substituted image. Unchanged-answer rate: its complement, an agreement statistic. Settling layer: first layer from which the lens-projected answer no longer changes, a property of the projection. Never diverged: share of pairs whose projected answers never differ at any layer. n/r: not reported, unlike a measured zero. Answer space and readout: MIMIC-CXR, yes/no from the first answer-token logits with the report in context; OmniMedVQA, one of four option letters; VQA-RAD, SLAKE and ProbMed, yes/no from the final-token answer logits. Model revisions: MedGemma-27B 2d3e00e, MedGemma-4B 290cda5e, Qwen3.5-9B c2022362.

How the MIMIC-CXR cohort and the other case pools were paired and labelled. Case pools by release, which bound every later denominator (the MIMIC-CXR cohort is fixed in the main text): OmniMedVQA, swap-changed case mining, pool 127{,}995, 427 admissible pairs, 33 swap-changed cases; SLAKE, swap-changed case inventory, pool 186, 90 swap-changed cases, swap-changed rate 0.484; VQA-RAD, pool ceiling mining, 52 admissible pairs, leakage-rule ceiling 66; VQA-RAD, pool relaxation, image reuse, 52 admissible pairs, leakage-rule ceiling 66. The OmniMedVQA answer-key audit flags 104 of 427 cases (0.244) whose correct option's text appears in the question; original-image accuracy is 0.576 overall and 0.607 on the 323 unflagged cases, and adversarial relabelling of every flagged case would bound it between 0.459 and 0.703; flip and unchanged-answer rates never read the key. On VQA-RAD, the automatic unchanged-answer label and the answer key's own image-decisive marking agree on 0.726 of 266 question-image rows (92 questions), Cohen's kappa 0.430.

\section{Robustness of the main result}
The headline design re-run with changed prompt, input order, resolution, report section, decoding and image content; none of it enters the main-text estimates. Swap tier, flip rate pooled and over pairs whose substitute shares the source or organ (hard, as in the main experiment) or not (easy): OmniMedVQA, same-source against cross-source, 427 pairs: 0.068 pooled, 0.068 hard; VQA-RAD, same-organ against different-organ, 52 pairs: 0.615 pooled, 0.526 hard, 0.857 easy. Modality ablation on the MIMIC-CXR cohort, one question per case (its labelled finding), accuracy against the report-derived label as in Table~\ref{tab:ablate}: one input dropped (3{,}199 cases): 0.931 full prompt, 0.913 report plus question, 0.773 image plus question, 0.846 question alone; the image adds 0.018; one input dropped, re-run (3{,}199 cases): 0.931 full prompt, 0.913 report plus question, 0.773 image plus question, 0.846 question alone; the image adds 0.017. Response-bias floor of the unchanged-answer rate on MIMIC-CXR, the agreement a responder answering from its marginal alone would reach: over 7 per-finding evaluations, each over that finding's 3{,}199 pairs, the floor runs from 0.500 to 0.919.

\begin{table}[H]
\centering\small
\setlength{\tabcolsep}{3.5pt}
\adjustbox{max width=\textwidth}{%
\begin{tabular}{llrrrrrr}
\toprule
Manipulation & Arm & $n$ & Flip rate & Flips & Settling layer & Never div. & Still apart \\
\midrule
prompt wording & clinical wording, yes/no order & 185 & 1.62\% [0.55, 4.66] & 3 & 59.0 & 64.32\% & n/r \\
prompt wording & direct wording, no/yes order & 185 & 4.32\% [2.21, 8.30] & 8 & 62.0 & 64.86\% & n/r \\
prompt wording & direct wording, yes/no order & 185 & 1.62\% [0.55, 4.66] & 3 & 59.0 & 63.24\% & n/r \\
prompt wording & impression wording, yes/no order & 185 & 1.08\% [0.30, 3.86] & 2 & 59.0 & 61.08\% & n/r \\
modality order & image before text & 200 & 3.00\% & 6 & 59.0 & 25.50\% & 48 \\
modality order & text before image & 200 & 16.50\% & 33 & 62.0 & 2.50\% & 122 \\
resolution & downsampled 2x & 200 & 2.00\% & 4 & n/r & n/r & 47 \\
resolution & downsampled 4x & 200 & 2.00\% & 4 & n/r & n/r & 51 \\
resolution & native resolution & 200 & 3.00\% & 6 & n/r & n/r & 48 \\
report section & findings and impression & 200 & 3.00\% & 6 & 57.0 & n/r & 48 \\
report section & findings only & 200 & 2.50\% & 5 & 59.0 & n/r & 51 \\
report section & impression only & 200 & 25.50\% & 51 & 58.0 & n/r & 97 \\
\bottomrule
\end{tabular}}
\caption{MIMIC-CXR under another prompt wording, modality order, resolution or report section; Settling layer is the median over pairs, Still apart the pairs whose projections differ at the final layer. The main experiment: image first, native, both sections. Brackets: Wilson 95\% intervals; unbracketed cells have none. These are 200-case pilots and are superseded on the full cohort: modality order is rerun on all 3{,}199 cases in Table~\ref{tab:designs} (2.84\% text first against 1.41\% image first), and impression only is rerun on the 2{,}394 cases that carry an impression, 4.26\% [3.34, 5.31], and on the 1{,}702 cases carrying both sections in Appendix Table~\ref{tab:textctl}. Read the 25.50\% here as a 200-case pilot, not as the paper's estimate.}
\label{tab:sup-mimicarms}
\end{table}

\begin{table}[H]
\centering\small
\setlength{\tabcolsep}{3.5pt}
\adjustbox{max width=\textwidth}{%
\begin{tabular}{llrrrrr}
\toprule
Dataset & Arm & $n$ & Samples & Swap flip, greedy & Floor at $T{=}0$ & Floor at $T{=}1$ \\
\midrule
MIMIC-CXR & repeated decoding, 40-case pilot & 40 & 5 & 0.00\% & 0.00\% & 42.50\% \\
OmniMedVQA & repeated decoding, 40-case pilot & 40 & 10 & 5.00\% & 0.00\% & 20.00\% \\
VQA-RAD & repeated decoding, full pool & 52 & 10 & 61.54\% [48.08, 75.00] & 0.00\% [0.00, 0.00] & 21.15\% [11.54, 32.69] \\
\bottomrule
\end{tabular}}
\caption{Decoding-noise floor: answer-change rate between repeats at temperature 0 and 1.0, beside the greedy swap flip rate. Runs decode greedily, so the temperature-0 floor is zero by construction; the temperature-1 floor is a same-input control. Brackets: 95\% bootstrap intervals; unbracketed cells have none.}
\label{tab:sup-noise}
\end{table}

\begin{table}[H]
\centering\small
\setlength{\tabcolsep}{3.5pt}
\adjustbox{max width=\textwidth}{%
\begin{tabular}{llrrcrr}
\toprule
Dataset & Arm & $n$ & Agreement & 95\% CI & Agree & Disagree \\
\midrule
MIMIC-CXR & blank image & 200 & 97.50\% & n/r & 195 & 5 \\
SLAKE & blank image & 186 & 55.38\% & [48.39, 62.37] & 103 & 83 \\
VQA-RAD & blank image & 52 & 61.54\% & n/r & n/r & n/r \\
\bottomrule
\end{tabular}}
\caption{Blank-image condition (image tokens present, content blank): share of cases whose answer with the blank image equals the answer with the real image. Agreement means only that the answer did not change. 95\% CI: patient-level bootstrap, 10{,}000 draws. n/r: not recorded for that run.}
\label{tab:sup-blank}
\end{table}

\begin{table}[H]
\centering\small
\setlength{\tabcolsep}{3.5pt}
\adjustbox{max width=\textwidth}{%
\begin{tabular}{llrrrrr}
\toprule
Dataset & Arm & $n$ & Clean acc. & Blur, max & Noise, max & Occlusion, max \\
\midrule
OmniMedVQA & graded blur, noise and occlusion & 427 & 0.576 [0.529, 0.623] & 7.49\% [5.15, 10.07] & 6.79\% [4.45, 9.37] & 9.84\% [7.03, 12.65] \\
SLAKE & graded blur, noise and occlusion & 186 & 0.710 [0.645, 0.774] & 24.19\% [18.28, 30.65] & 17.20\% [11.83, 22.58] & 44.62\% [37.63, 51.61] \\
VQA-RAD & graded corruption & 52 & 0.731 [0.615, 0.846] & 28.85\% & 34.62\% & 38.46\% [25.00, 51.92] \\
\bottomrule
\end{tabular}}
\caption{Graded image corruption: share of cases whose answer changes at the strongest blur, noise and occlusion, beside clean-image accuracy. VQA-RAD: corrected readout 0.731 supersedes the earlier double-normed 0.654. Brackets: 95\% intervals (OmniMedVQA, bootstrap, 2{,}000 draws; SLAKE and VQA-RAD, bootstrap); unbracketed cells have none.}
\label{tab:sup-corruption}
\end{table}

\begin{table}[H]
\centering\small
\setlength{\tabcolsep}{3.5pt}
\adjustbox{max width=\textwidth}{%
\begin{tabular}{llrrrrr}
\toprule
Dataset & Arm & $n$ & Lowest dose & Highest dose & Trend $\rho$ & Trend $p$ \\
\midrule
MIMIC-CXR & report length, text-side confound & 185 & 2.13\% & 0.00\% & $-0.057$ & 0.443 \\
MIMIC-CXR & swap severity, label Hamming distance & 185 & 0.00\% [0.00, 29.91] & 2.27\% [0.63, 7.91] & 0.044 & 0.549 \\
SLAKE & swap severity, pixel distance & 186 & 47.83\% [32.61, 63.04] & 62.50\% [47.92, 75.00] & 0.115 & 0.115 \\
VQA-RAD & swap severity, divergence tertiles & 52 & 12.50\% & 100.00\% & n/r & n/r \\
\bottomrule
\end{tabular}}
\caption{Swap-severity dose-response: flip rate in the lowest and highest bucket (each over its pairs), Spearman's rank correlation with severity across buckets and its two-sided $p$; Arm names the severity measure. Brackets: 95\% intervals (MIMIC-CXR, Wilson; SLAKE, bootstrap); unbracketed cells have none.}
\label{tab:sup-dose}
\end{table}

\begin{table}[H]
\centering\small
\setlength{\tabcolsep}{3.5pt}
\adjustbox{max width=\textwidth}{%
\begin{tabular}{lrrrr}
\toprule
Finding & Observed & Chance-expected & Raw excess & Kappa \\
\midrule
Atelectasis & 0.951 & 0.698 & 0.254 & 0.839 [0.807, 0.867] \\
Cardiomegaly & 0.936 & 0.500 & 0.435 & 0.871 [0.845, 0.896] \\
Consolidation & 0.931 & 0.559 & 0.373 & 0.845 [0.818, 0.868] \\
Edema & 0.963 & 0.514 & 0.449 & 0.923 [0.906, 0.940] \\
Enlarged Cardiomediastinum & 0.951 & 0.505 & 0.446 & 0.901 [0.883, 0.919] \\
Fracture & 0.988 & 0.861 & 0.126 & 0.910 [0.866, 0.948] \\
Lung Lesion & 0.914 & 0.514 & 0.401 & 0.824 [0.800, 0.846] \\
Lung Opacity & 0.960 & 0.763 & 0.197 & 0.832 [0.793, 0.865] \\
Pleural Effusion & 0.940 & 0.504 & 0.436 & 0.880 [0.860, 0.898] \\
Pleural Other & 0.956 & 0.500 & 0.455 & 0.911 [0.894, 0.927] \\
Pneumonia & 0.950 & 0.583 & 0.367 & 0.880 [0.856, 0.903] \\
Pneumothorax & 0.992 & 0.919 & 0.073 & 0.899 [0.837, 0.943] \\
Support Devices & 0.946 & 0.914 & 0.032 & 0.370 [0.271, 0.451] \\
No Finding & 0.964 & 0.501 & 0.463 & 0.927 [0.913, 0.941] \\
\bottomrule
\end{tabular}}
\caption{Cohen's kappa with condition-specific marginals, MIMIC-CXR, 14 findings: Observed is the finding's unchanged-answer rate, Chance-expected is $p_c p_d + (1-p_c)(1-p_d)$ from the two conditions' yes-rates, Kappa the excess over $1 - \text{expected}$. Per row: 3{,}199 pairs per finding. Brackets on kappa: 95\% patient-clustered percentile bootstrap over the 293 patients, 2{,}000 draws at seed 0, resampling the agreement and both yes-rate indicators jointly so each draw rebuilds the marginals kappa depends on.}
\label{tab:sup-kappa}
\end{table}

\section{Other datasets and models}
These experiments evaluate the swap protocol on datasets without reports and through a second lens; they are not part of it. Accuracy under the swap on OmniMedVQA: all 427 of 427 swaps change the correct answer, so accuracy is scored against each condition's ground truth (chance 0.25): 0.579 on the original image, 0.457 on the substituted one, a drop of 0.122 pooled, $-0.003$ on anatomy identification and 0.654 on disease diagnosis. Text-side symmetry, share of cases whose answer changes: on OmniMedVQA (427 cases), 0.232 for a question swap with the image unchanged, 0.021 for an option swap, 0.068 for an image swap and 0.750 for a uniform option reroll; on SLAKE (186 cases), 0.419 for a question swap with the image unchanged, 0.484 for an image swap and 0.500 for a uniform option reroll; on VQA-RAD (52 cases), 0.269 for a question swap with the image unchanged, 0.615 for an image swap. Open-ended generation, a frozen rule mapping the free-text answer to an option: on OmniMedVQA, 427 of 427 answers mapped and 426 also cleared the 0.500 first-token probability floor (chance 0.250); on SLAKE, 111 of 186 answers mapped and 0 also cleared the 0.500 first-token probability floor; on VQA-RAD, 4 of 52 answers mapped and 0 also cleared the 0.500 first-token probability floor.

\begin{table}[H]
\centering\small
\setlength{\tabcolsep}{3.5pt}
\adjustbox{max width=\textwidth}{%
\begin{tabular}{llrrrrrr}
\toprule
Dataset & Arm & $n$ & Both inputs & Question only & Image, no question & Neither & Image adds \\
\midrule
OmniMedVQA & one input dropped, or both & 427 & 0.576 [0.529, 0.623] & 0.543 [0.496, 0.590] & 0.508 [0.461, 0.555] & 0.515 [0.468, 0.564] & $+0.033$ [0.002, 0.061] \\
SLAKE & one input dropped, or both & 186 & 0.710 [0.645, 0.774] & 0.500 [0.430, 0.570] & 0.500 [0.430, 0.570] & 0.500 [0.430, 0.570] & $+0.210$ [0.118, 0.301] \\
VQA-RAD & inputs dropped & 52 & 0.731 [0.615, 0.846] & 0.538 [0.404, 0.673] & 0.500 & 0.500 & $+0.192$ [0.038, 0.346] \\
\bottomrule
\end{tabular}}
\caption{Modality ablation without a report: accuracy against the answer key with both inputs, question only, image under a neutral prompt, or neither; chance 0.25 on OmniMedVQA, 0.5 elsewhere. VQA-RAD: corrected readout 0.731 supersedes the earlier 0.654. Brackets: 95\% bootstrap intervals; unbracketed cells have none.}
\label{tab:sup-ablationnr}
\end{table}

\begin{table}[H]
\centering\small
\setlength{\tabcolsep}{3.5pt}
\adjustbox{max width=\textwidth}{%
\begin{tabular}{llrrrr}
\toprule
Dataset & Arm & $n$ & Concordant & Discordant & Flip rate \\
\midrule
SLAKE & logit lens, full pool & 186 & 0.710 [0.645, 0.774] & 0.710 [0.645, 0.774] & 48.39\% [41.40, 55.91] \\
SLAKE & tuned lens, full pool & 186 & 0.661 [0.591, 0.731] & 0.661 [0.591, 0.726] & 34.41\% [27.96, 41.40] \\
VQA-RAD & logit lens & 52 & 0.731 & 0.731 & 61.54\% \\
VQA-RAD & tuned lens, 2{,}000 steps & 52 & 0.673 & 0.673 & 34.62\% \\
\bottomrule
\end{tabular}}
\caption{Readout comparison: concordant and discordant accuracy and flip rate of the answer projected at the final decoder layer by the output head (logit lens) or trained translator (tuned lens), not the model's output. Brackets: 95\% bootstrap intervals, 10{,}000 draws; unbracketed cells have none.}
\label{tab:sup-lenseval}
\end{table}

\begin{table}[H]
\centering\small
\setlength{\tabcolsep}{3.5pt}
\adjustbox{max width=\textwidth}{%
\begin{tabular}{llrrrr}
\toprule
Dataset & Half & $n$ & Unchanged & Flip & Never div. \\
\midrule
OmniMedVQA & development half & 175 & 97.71\% [95.43, 99.43] & 2.29\% [0.57, 4.57] & 0.57\% \\
OmniMedVQA & replication half & 252 & 90.08\% [86.51, 93.65] & 9.92\% [6.35, 13.49] & 7.54\% \\
SLAKE & development half & 92 & 50.00\% [39.13, 59.78] & 50.00\% [40.22, 60.87] & 13.04\% \\
SLAKE & replication half & 94 & 53.19\% [43.62, 62.77] & 46.81\% [37.23, 56.38] & 10.64\% [5.32, 17.02] \\
\bottomrule
\end{tabular}}
\caption{Frozen-protocol replication without a report: split by swap group fixed before any outcome was seen, analysis fixed on the development half, replication half run once. Unchanged and flip rates are complements. Brackets: 95\% bootstrap intervals; unbracketed cells have none.}
\label{tab:sup-frozen}
\end{table}

\section{Exploratory mechanistic analyses}
Layer-wise projections and interventions, descriptive of the lens projection; the main text draws no mechanistic claim from them. Activation patching over flip cases only: OmniMedVQA, patching with no-op control, 29 flip cases of 427 over 62 layers: patching restores the original answer most often at layer 60 and on more than half the flip cases first at layer 44; never-diverged share 0.047; VQA-RAD, activation patching, 32 flip cases of 52 over 62 layers: patching restores the original answer most often at layer 58 and on more than half the flip cases first at layer 31; never-diverged share 0.038. Tuned-lens training for MedGemma-27B, one run each, not used by the main experiment: WikiText, 100 steps, best validation KL 829.1 at step 100, 100 steps, 208 s wall time; WikiText-103, 1{,}000 steps, best validation KL 233.4 at step 900, 1{,}000 steps, 1{,}270 s wall time; WikiText-103, 2{,}000 steps, best validation KL 195.4 at step 1{,}700, 2{,}000 steps, 2{,}492 s wall time, final validation KL 197.6; seed reproducibility check, 60 steps. Probes, a linear classifier of swap-induced answer change on the residual at the stated layer: OmniMedVQA, cross-model probe, LLaVA-1.5-7B, 22 swap-changed cases of 427, out-of-fold balanced accuracy 0.572 against a matched shuffled null of 0.474; OmniMedVQA, leave-one-modality-out transfer, layer 45, 29 swap-changed cases of 427, 0 transfer folds above the null; VQA-RAD, leave-one-category-out transfer, layer 20, 32 swap-changed cases of 52; VQA-RAD, probe curve and stability, 32 swap-changed cases of 52, out-of-fold balanced accuracy 0.516 against a matched shuffled null of 0.444. Saliency baselines against the probe, area under the risk-coverage curve (lower is better) for abstention: OmniMedVQA (layer 45, 427 cases), 0.354 with the probe's score, 0.450 with image-token attention mass and 0.437 with the input gradient; VQA-RAD (layer 20, 52 cases), 0.199 with the probe's score, 0.210 with image-token attention mass and 0.320 with the input gradient.

\begin{table}[H]
\centering\small
\setlength{\tabcolsep}{3.5pt}
\adjustbox{max width=\textwidth}{%
\begin{tabular}{llrrrrr}
\toprule
Dataset & Arm & $n$ & Flip rate & Unchanged & Peak div. & Layers \\
\midrule
MIMIC-CXR & test split, 300 cases & 300 & 2.00\% & 98.00\% & 0.078 & 62 \\
MIMIC-CXR & test split, 976 cases & 976 & 1.43\% & 98.57\% & 0.076 & 62 \\
MIMIC-CXR & train split, 300 cases & 300 & 0.67\% & 99.33\% & 0.077 & 62 \\
MIMIC-CXR & train split, 976 cases & 976 & 0.92\% & 99.08\% & 0.074 & 62 \\
MIMIC-CXR & validation split, 300 cases & 300 & 1.33\% & 98.67\% & 0.072 & 62 \\
MIMIC-CXR & validation split, 880 cases & 880 & 1.02\% & 98.98\% & 0.076 & 62 \\
ProbMed & cross-image, same question & 2{,}634 & 39.86\% & 60.14\% & 0.633 & 62 \\
ProbMed & full test set, 27B & 6{,}158 & 68.87\% & 31.13\% & 1.000 & 62 \\
ProbMed & full test set, 4B & 6{,}158 & 71.65\% & 28.35\% & 1.000 & 34 \\
\bottomrule
\end{tabular}}
\caption{Settling depth by split: Peak div. is the median over pairs of the peak per-layer divergence between the two images' projected answer distributions, Layers the decoder depth. The MIMIC-CXR rows are earlier, smaller samples than Table~\ref{tab:head}. No intervals were computed.}
\label{tab:sup-settling}
\end{table}

\begin{table}[H]
\centering\small
\setlength{\tabcolsep}{3.5pt}
\adjustbox{max width=\textwidth}{%
\begin{tabular}{llrrrrrr}
\toprule
Dataset & Arm & $n$ & Cells & Computable & Significant & Median, min & Median, max \\
\midrule
MIMIC-CXR & settling-definition grid & 3{,}199 & 24 & n/r & n/r & 58.0 & 62.0 \\
OmniMedVQA & definition grid, unchanged-answer cases & 427 & 24 & 15 & 7 & 42.0 & 59.5 \\
VQA-RAD & definition grid, unchanged-answer cases & 52 & 24 & 5 & 0 & 35.0 & 61.0 \\
\bottomrule
\end{tabular}}
\caption{Settling-definition sensitivity over a divergence-metric by threshold grid: grid cells, computable cells, cells where the two conditions' ordering stayed significant, and the range of the median settling layer over computable cells. No intervals were computed.}
\label{tab:sup-defsens}
\end{table}

\begin{table}[H]
\centering\small
\setlength{\tabcolsep}{3.5pt}
\adjustbox{max width=\textwidth}{%
\begin{tabular}{llrrrrrr}
\toprule
Dataset & Arm & $n$ & Layers & Peak layer & Peak-layer change & All-layer change & Random-key change \\
\midrule
OmniMedVQA & image attention masked, layer sweep & 427 & 62 & 8 & 8.20\% [5.62, 11.01] & 7.96\% [5.62, 10.77] & 83.61\% [80.09, 86.89] \\
SLAKE & image attention masked, layer sweep & 186 & 62 & 0 & 4.30\% [1.61, 7.53] & 36.56\% [29.57, 43.55] & n/r \\
VQA-RAD & image attention masked, layer sweep & 52 & 62 & 23 & 34.62\% [21.15, 48.08] & 48.08\% [34.62, 61.54] & 51.92\% [38.46, 65.38] \\
\bottomrule
\end{tabular}}
\caption{Attention knockout, layer-swept: share of cases whose answer changes with text-to-image attention masked at every layer, at the most effective layer only, or at random key positions of matched size. Nothing is localised. Brackets: 95\% bootstrap intervals; unbracketed cells have none.}
\label{tab:sup-knockout}
\end{table}

\begin{table}[H]
\centering\small
\setlength{\tabcolsep}{3.5pt}
\adjustbox{max width=\textwidth}{%
\begin{tabular}{llrrrrrr}
\toprule
Dataset & Arm & $n$ & Layer & Cases steered & Flip & Selective flip & Random flip \\
\midrule
OmniMedVQA & positive control, answer axis & 427 & 45 & 427 & 18.27\% & n/r & 21.78\% \\
SLAKE & positive control, answer axis & 186 & 61 & 186 & 56.99\% [50.00, 63.98] & 6.23\% & n/r \\
VQA-RAD & positive control, answer axis & 52 & 40 & 52 & 50.00\% & n/r & 11.54\% \\
MIMIC-CXR & non-linear direction, layer 20 & n/r & 20 & n/r & n/r & 4.92\% & n/r \\
MIMIC-CXR & per-layer efficacy sweep & 200 & 48 & 187 & 41.18\% [34.22, 48.13] & 38.02\% & 19.79\% [14.44, 25.67] \\
OmniMedVQA & layer sweep, sub-saturation scales & 427 & 45 & 398 & 2.01\% & 1.60\% & 1.51\% \\
SLAKE & steering l61 relative hi & 186 & 61 & 98 & 38.78\% & 0.00\% & 38.78\% \\
VQA-RAD & difference-of-means, random null & 52 & 40 & 20 & 10.00\% & 10.00\% & 10.00\% \\
VQA-RAD & per-layer efficacy sweep & 52 & 8 & 20 & 60.00\% [40.00, 80.00] & 7.37\% & 70.00\% [50.00, 90.00] \\
\bottomrule
\end{tabular}}
\caption{Steering, positive controls (all cases steered) and sweeps (unchanged-answer cases), best cell: flip, share whose steered readout differs from the unsteered; selective flip, minus new errors among correct cases; random flip, largest under a magnitude-matched random direction. Brackets: 95\% bootstrap intervals; unbracketed cells have none.}
\label{tab:sup-steer}
\end{table}

\begin{table}[H]
\centering\small
\setlength{\tabcolsep}{3.5pt}
\adjustbox{max width=\textwidth}{%
\begin{tabular}{llrrrrr}
\toprule
Dataset & Arm & $n$ & Probe layer & AURC probe & AURC max-softmax & AURC random \\
\midrule
OmniMedVQA & abstention regime sweep & 427 & 45 & 0.354 & 0.304 & 0.382 \\
SLAKE & AURC, held-out layer selection & 186 & 35 & 0.179 & 0.228 & 0.169 \\
VQA-RAD & AURC, swap-grouped folds & 52 & 20 & 0.199 & 0.134 & n/r \\
\bottomrule
\end{tabular}}
\caption{Abstention: area under the risk-coverage curve (lower is better) when the probe's score, the answer's maximum softmax or a random score decides which cases to abstain on; the probe is a linear classifier of swap-induced answer change. No intervals were computed.}
\label{tab:sup-aurc}
\end{table}

\section{Runs that did not yield an interpretable result}
Experiments whose records judge the result not interpretable: obsolete runs on an earlier cohort, and a readout that disagrees with the model's output. Reported for completeness; none is compared with the main text. Scale and finetuning on the MIMIC-CXR atelectasis cohort, one question per case, logit lens against the report-derived label, obsolete runs not comparable with Table~\ref{tab:head}: 4B against 27B, settling depth (3{,}199 cases), relative settling depth 0.735 for MedGemma-4B against 1.000 for MedGemma-27B under the report prompt; Gemma-3 against MedGemma, accuracy (3{,}199 cases), accuracy 0.433 for Gemma-3-27B against 0.415 for MedGemma-27B at layer 41 of 62, report condition unstated, majority baseline 0.777. The tuned-lens readout on the ProbMed test set, 6{,}158 pairs: flip rate 0.509, unchanged-answer rate 0.491, never-diverged share 0.008, for the answer projected by the text-trained translator at the final layer, which matches the model's output on 0.607 of cases; not interpretable.

\end{document}